\documentclass{article}

\usepackage[main, final]{neurips_2026}

\usepackage[utf8]{inputenc} %
\usepackage[T1]{fontenc}    %
\usepackage{url}            %
\usepackage{booktabs}       %
\usepackage{amsfonts}       %
\usepackage{nicefrac}       %
\usepackage{microtype}      %
\usepackage{xcolor}  %
\usepackage{tikz}
\usetikzlibrary{decorations.markings}

\definecolor{eggshell}{HTML}{F0EAD6} %
\definecolor{icmlViolet}{HTML}{4169e1} 
\definecolor{darkEggshell}{HTML}{837b6d} %
\definecolor{mydarkblue}{rgb}{0,0.08,0.45}

\usepackage[
    colorlinks=true, 
    linkcolor=darkEggshell,     %
    urlcolor=darkEggshell,      %
    citecolor=mydarkblue, %
    anchorcolor=eggshell
]{hyperref}

\usepackage{wrapfig}
\usepackage{tabularx}
\usepackage[export]{adjustbox}
\usepackage{color,soul}
\usepackage{xcolor}
\usepackage{enumitem}
\usepackage{setspace}
\usepackage{xspace}
\usepackage{longtable}
\usepackage{siunitx}
\usepackage{listings}
\usepackage{multibib}
\usepackage{seqsplit} 
\usepackage{natbib}
\usepackage{graphicx}

\usepackage{algorithm}
\usepackage{algpseudocode}

\usepackage[normalem]{ulem}  %

\newif\ifshowchanges
\showchangestrue   %

\ifshowchanges
  \newcommand{\del}[1]{{\color{red}\sout{#1}}}        %
\else
  
  \newcommand{\del}[1]{}
\fi

\usepackage{amsmath,amsfonts,bm}

\def\1{\bm{1}}

\DeclareMathAlphabet{\mathsfit}{\encodingdefault}{\sfdefault}{m}{sl}
\SetMathAlphabet{\mathsfit}{bold}{\encodingdefault}{\sfdefault}{bx}{n}

\newcommand{\R}{\mathbb{R}}

\usepackage[framemethod=TikZ]{mdframed}
\mdfdefinestyle{MyFrame}{%
    linecolor=black,
    outerlinewidth=0.5pt,
    roundcorner=1pt,
    innertopmargin=2pt, %
    innerbottommargin=2pt, %
    innerrightmargin=7pt,
    innerleftmargin=7pt,
    backgroundcolor=black!0!white}

\mdfdefinestyle{MyFrame2}{%
    linecolor=white,
    outerlinewidth=1pt,
    roundcorner=2pt,
    innertopmargin=10pt,
    innerbottommargin=0.5pt,
    innerrightmargin=10pt,
    innerleftmargin=10pt,
    backgroundcolor=black!3!white}

\mdfdefinestyle{MyFrameEq}{%
    linecolor=white,
    outerlinewidth=0pt,
    roundcorner=0pt,
    innertopmargin=0pt,
    innerbottommargin=0pt,
    innerrightmargin=7pt,
    innerleftmargin=7pt,
    backgroundcolor=black!3!white}

\mdfdefinestyle{FrameAdded}{%
    linecolor=black,
    outerlinewidth=0pt,
    roundcorner=0pt,
    innertopmargin=0pt,
    innerbottommargin=0pt,
    innerrightmargin=0pt,
    innerleftmargin=0pt,
    backgroundcolor=PastelGreen}

\definecolor{customwhite}{HTML}{FCFBF7}
\definecolor{customturq}{HTML}{1D9D79}
\definecolor{customorange}{HTML}{D96002} 
\definecolor{custombeige}{HTML}{d6c9b1} 

\definecolor{custompurple}{HTML}{AEADF0}
\definecolor{customwhite2}{HTML}{fbf9f4}
\definecolor{customblue}{HTML}{4D9DE0}

\usepackage{silence}
\usepackage[utf8]{inputenc}
\usepackage[T1]{fontenc}
\usepackage{tabularx}
\usepackage{lipsum}
\usepackage{url}
\usepackage{placeins}
\usepackage{booktabs}
\usepackage{soul}
\usepackage{amsfonts}
\usepackage{nicefrac}
\usepackage{microtype}
\usepackage{wrapfig}
\usepackage{caption}
\usepackage{amssymb}
\usepackage{pifont}

\usepackage{subcaption}
\usepackage{float}
\usepackage{rotating}
\usepackage{etoolbox}
\usepackage{xparse}
\usepackage{grffile}
\usepackage{amsmath}
\usepackage{xspace}
\usepackage{euscript}
\usepackage{enumitem}
\usepackage{mathtools}
\usepackage{tikz}
\usepackage{tablefootnote}
\usepackage[flushleft]{threeparttable}
\usepackage{multirow}
\usepackage[title]{appendix}
\usepackage{arydshln}
\usepackage{array, booktabs}
\usepackage{graphicx}
\usepackage{amsmath,amsthm}
\usepackage{dsfont}
\usepackage{balance}
\usepackage{multicol}
\usepackage{xcolor}
\usepackage{nameref}
\usepackage{pdfpages}
\usepackage{braket}
\usepackage[normalem]{ulem}
\usepackage{bbding}
\usepackage{xfrac}
\usepackage[usestackEOL]{stackengine}
\usepackage{indentfirst}
\usepackage{csquotes}
\usepackage{pgf}
\usepackage{varwidth}
\usepackage{verbatimbox}
\usepackage{verbatim}
\usepackage{bm}
\usepackage{anyfontsize}

\usepackage{empheq}
\definecolor{myblue}{rgb}{.8, .8, 1}

\definecolor{pastelblue}{RGB}{76,113,175}
\definecolor{pastelgreen}{RGB}{144,238,144}
\definecolor{pastelred}{RGB}{196,78,82}
\definecolor{pastelgrey}{RGB}{230,230,230}
\definecolor{pastelbeige}{RGB}{243,236,221}
\definecolor{pastelpurple}{RGB}{154,139,192}
\definecolor{salmon}{RGB}{250, 128, 114}
\definecolor{darkgreen}{rgb}{0,0.6,0}
\definecolor{darkred}{rgb}{0.5,0,0}
\definecolor{verylightgreen}{HTML}{F6FFF9}
\definecolor{verylightred}{HTML}{FFF4F3}
\definecolor{verylightgray}{HTML}{F4F6F6}
\definecolor{babyblueeyes}{rgb}{0.63, 0.79, 0.95}
\definecolor{lightpink}{rgb}{1.00, 0.714, 0.757}

\usepackage{soul}

\definecolor{some_color}{HTML}{86CFDA}
\colorlet{PastelGreen}{some_color!50!white}
\sethlcolor{PastelGreen}

\usepackage{tikzpeople}
\usetikzlibrary{shapes,decorations,arrows,calc,arrows.meta,fit,positioning}

\tikzset{
    -Latex,auto,node distance =1 cm and 1 cm,semithick,
    state/.style ={ellipse, draw, minimum width = 0.7 cm},
    point/.style = {circle, draw, inner sep=0.04cm,fill,node contents={}},
    bidirected/.style={Latex-Latex,dashed},
    el/.style = {inner sep=2pt, align=left, sloped}
}

\makeatletter
\def\thmt@refnamewithcomma #1#2#3,#4,#5\@nil{%
	\@xa\def\csname\thmt@envname #1utorefname\endcsname{#3}%
	\ifcsname #2refname\endcsname
	\csname #2refname\expandafter\endcsname\expandafter{\thmt@envname}{#3}{#4}%
	\fi}
\makeatother

\newcommand*\dbar[1]{\overline{\overline{\lower0.2ex\hbox{$#1$}}}}

\def\cT{{\mathcal{T}}}

\def\cX{{\mathcal{X}}}
\def\cY{{\mathcal{Y}}}

\DeclareFontFamily{U}{BOONDOX-calo}{\skewchar\font=45 }
\DeclareFontShape{U}{BOONDOX-calo}{m}{n}{
  <-> s*[1.05] BOONDOX-r-calo}{}
\DeclareFontShape{U}{BOONDOX-calo}{b}{n}{
  <-> s*[1.05] BOONDOX-b-calo}{}
\DeclareMathAlphabet{\mathcalb}{U}{BOONDOX-calo}{m}{n}
\SetMathAlphabet{\mathcalb}{bold}{U}{BOONDOX-calo}{b}{n}
\DeclareMathAlphabet{\mathbcalb}{U}{BOONDOX-calo}{b}{n}

\usepackage{cancel}

\renewcommand{\paragraph}[1]{{\noindent \textbf{#1.}}}

\usepackage{listings}

\usepackage{tikz}
\usetikzlibrary{calc}

\definecolor{SourceGrey}{HTML}{E0E0E0}
\definecolor{MFMPink}{HTML}{E6CECF}
\definecolor{DeepBlue}{HTML}{297aff}
\definecolor{TargetGreen}{HTML}{1D9D79}
\definecolor{StaticPurple}{HTML}{8E44AD} %

\definecolor{method_blue}{HTML}{2979FF}   %
\definecolor{op_pink}{HTML}{B08B8C}       %
\definecolor{pastel_green}{HTML}{2F8C76}  %
\definecolor{text_black}{HTML}{000000}    %

\newcommand{\tfm}{\texttt{M2M-TFM}\xspace}
\newcommand{\edmap}{\texttt{M2M-ED}\xspace}

\usepackage{fontawesome5} %

\makeatletter
\def\blfootnote{\xdef\@thefnmark{}\@footnotetext}
\makeatother

\usetikzlibrary{decorations.markings, arrows.meta}

\let\originalleft\left
\let\originalright\right
\renewcommand{\left}{\mathopen{}\mathclose\bgroup\originalleft}
\renewcommand{\right}{\aftergroup\egroup\originalright}

\definecolor{antiquefuchsia}{rgb}{0.57, 0.36, 0.51}
\definecolor{amethyst}{rgb}{0.6, 0.4, 0.8}

\newcommand{\var}{{\rm I\kern-.3em D}}

\newtheorem*{theorem*}{Theorem}
\newtheorem*{proposition*}{Proposition}
\newtheorem*{example*}{Example}
\DeclareMathSymbol{\shortminus}{\mathbin}{AMSa}{"39}

\definecolor{pastel_purple}{HTML}{756FB3}

\colorlet{PastelPurpleLight}{pastel_purple!15!white}
\colorlet{PastelGreenLight}{pastel_green!15!white}
\sethlcolor{PastelPurpleLight}
\sethlcolor{PastelGreenLight}

\makeatletter
\let\save@mathaccent\mathaccent
\newcommand*\if@single[3]{%
  \setbox0\hbox{${\mathaccent"0362{#1}}^H$}%
  \setbox2\hbox{${\mathaccent"0362{\kern0pt#1}}^H$}%
  \ifdim\ht0=\ht2 #3\else #2\fi
  }
\newcommand*\rel@kern[1]{\kern#1\dimexpr\macc@kerna}
\newcommand*\widebar[1]{\@ifnextchar^{{\wide@bar{#1}{0}}}{\wide@bar{#1}{1}}}
\newcommand*\wide@bar[2]{\if@single{#1}{\wide@bar@{#1}{#2}{1}}{\wide@bar@{#1}{#2}{2}}}
\newcommand*\wide@bar@[3]{%
  \begingroup
  \def\mathaccent##1##2{%
    \let\mathaccent\save@mathaccent
    \if#32 \let\macc@nucleus\first@char \fi
    \setbox\z@\hbox{$\macc@style{\macc@nucleus}_{}$}%
    \setbox\tw@\hbox{$\macc@style{\macc@nucleus}{}_{}$}%
    \dimen@\wd\tw@
    \advance\dimen@-\wd\z@
    \divide\dimen@ 3
    \@tempdima\wd\tw@
    \advance\@tempdima-\scriptspace
    \divide\@tempdima 10
    \advance\dimen@-\@tempdima
    \ifdim\dimen@>\z@ \dimen@0pt\fi
    \rel@kern{0.6}\kern-\dimen@
    \if#31
      \overline{\rel@kern{-0.6}\kern\dimen@\macc@nucleus\rel@kern{0.4}\kern\dimen@}%
      \advance\dimen@0.4\dimexpr\macc@kerna
      \let\final@kern#2%
      \ifdim\dimen@<\z@ \let\final@kern1\fi
      \if\final@kern1 \kern-\dimen@\fi
    \else
      \overline{\rel@kern{-0.6}\kern\dimen@#1}%
    \fi
  }%
  \macc@depth\@ne
  \let\math@bgroup\@empty \let\math@egroup\macc@set@skewchar
  \mathsurround\z@ \frozen@everymath{\mathgroup\macc@group\relax}%
  \macc@set@skewchar\relax
  \let\mathaccentV\macc@nested@a
  \if#31
    \macc@nested@a\relax111{#1}%
  \else
    \def\gobble@till@marker##1\endmarker{}%
    \futurelet\first@char\gobble@till@marker#1\endmarker
    \ifcat\noexpand\first@char A\else
      \def\first@char{}%
    \fi
    \macc@nested@a\relax111{\first@char}%
  \fi
  \endgroup
}
\makeatother

\usepackage[capitalize,noabbrev]{cleveref}

\title{Measure-to-measure Regression with Transformers}

\author{
  \textbf{Matthew Vandergrift}$^{1}$ \qquad \textbf{Martha White}$^{1,2}$ \qquad \textbf{Yury Polyanskiy}$^{3}$ \\ 
  \textbf{Philippe Rigollet}$^{3}$ \qquad \textbf{Lazar Atanackovic}$^{1,2,4}$\thanks{Correspondence to: \url{atanacko@ualberta.ca}} \\
  \\
  $^1$University of Alberta, Alberta Machine Intelligence Institute, $^2$Canada CIFAR AI Chair \\ 
  $^3$MIT, $^4$The Broad Institute of MIT and Harvard }

\begin{document}

\maketitle

\vspace{-1.5em}
\begin{abstract}
\vspace{-0.5em}
Many learning problems require predicting how populations evolve under an unknown transformation. A natural representation for such populations is a probability measure, with point clouds as a key example. In this work, we study the \textbf{\textit{measure-to-measure} (M2M) regression} problem, in which one seeks to learn a map between probability measures from a finite collection of observed input--output measure pairs. In contrast to classical regression, where individual samples are transformed independently, M2M regression treats entire distributions as the data points. This perspective is vital in certain scientific applications, for example, cellular and molecular biology, where cells are known to evolve not as independent data points but as a collection. However, few existing approaches address the problem of M2M regression with sufficient expressivity and scalability. We present a formalization of nonlinear M2M regression and introduce two easy-to-use, expressive, and scalable approaches to learn such operators: transformers as \textbf{\textit{static}} M2M maps and transformers as \textbf{\textit{dynamic}} M2M velocity fields. Our approach leverages the natural measure-dependent and mean-field structure of transformers to learn nonlinear M2M maps on the space of probability distributions. We illustrate the effectiveness of our proposed methods to generalize to unseen measures on synthetic experiments, interacting particle systems, and a large-scale patient-derived organoid dataset for predicting treatment response in colorectal cancer.

\end{abstract}

\vspace{-1em}
\section{Introduction}
\label{sec:intro}
\vspace{-0.5em}

\begin{wrapfigure}{r}{0.5\textwidth}
    \vspace{-7.8em} 
    \includegraphics[width=0.5\columnwidth]{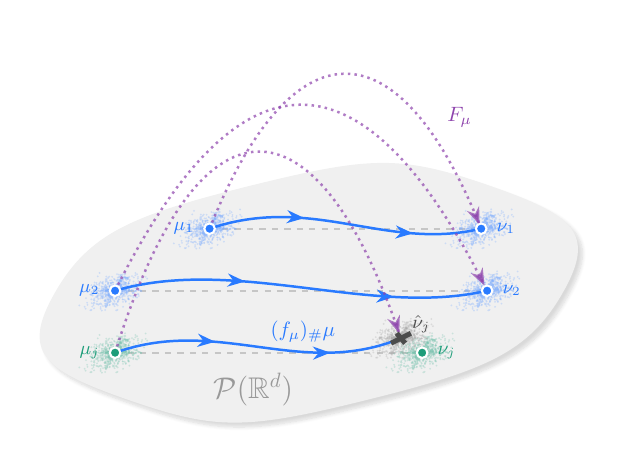}
    \vspace{-2.15em} 
    \caption{
    \textbf{Illustration of the \textit{measure-to-measure} (M2M) regression problem.} 
    In M2M regression, we treat entire distributions/measures as paired data points (e.g., $(\mu_i, \nu_i) \in \mathcal{P}(\mathbb{R}^d) \times \mathcal{P}(\mathbb{R}^d)$) to learn a continuous mapping between them. 
    Leveraging the natural measure-dependent structure of transformers, we introduce two classes of methods for learning nonlinear M2M operators: \textcolor{StaticPurple}{\textit{static}} one-step pushforward maps \textcolor{StaticPurple}{\smash{$F: \mu \mapsto (F_{\mu})_\# \mu$}} and \textcolor{DeepBlue}{\textit{dynamic}} flow  maps \textcolor{DeepBlue}{\smash{$f: \mu \mapsto (f_{\mu})_{\#}\mu$}} obtained by integrating an ODE on $\R^d$. 
    These learned operators can robustly generalize to \textit{unseen} test measures (predict \smash{$\hat{\nu}_j$} given novel input \textcolor{TargetGreen}{\smash{$\mu_j$}}).
    }
    \label{fig:m2m_tfm}
    \vspace{-19pt} %
\end{wrapfigure} 

Many modern datasets can be described as collections of distributions rather than individual vectors: bags of images, collections of time series, and sets of point clouds. As a result, a natural question arises: \textit{can we learn to model unknown transformations on the space of distributions?}  This is of particular interest in the natural sciences, where particles of physical systems are known to act as a collective and not independently. A motivating example arises from the advent of high-throughput single-cell perturbation screens \citep{dixit2016perturb, zapatero2023trellis, peidli2024scperturb, zhang2025tahoe, huang2025x}, which allow individuals' responses to stimuli to be measured by point clouds of cells in \textit{control} and \textit{treated} conditions.  Here, an unknown function maps \emph{pre-treatment} populations of cells to corresponding \emph{post-treatment} populations, and the modeling objective is to recover the ground-truth transformation \citep{Bunne2023NeuralOT, atanackovic2024meta, adduri2025predicting, wei2025benchmarking}. Since cells communicate and evolve as a collective rather than independently \citep{su2024cell, armingol2021deciphering, goodenough2009gap}, faithfully capturing their complex dynamics is a core modeling challenge, and one that calls for computational methods that explicitly account for such distribution-dependent behavior.

\vspace{-0.3em}
\paragraph{Measure-to-measure (M2M) regression} Modeling such transformations naturally corresponds to learning functions between probability measures on $\mathcal{P}(\R^d)$ (see \cref{fig:m2m_tfm}). Early work along these lines on \emph{distribution-to-distribution regression} relied on kernel methods \citep{oliva2013distribution}, while subsequent work on distribution regression has largely focused on settings in which either the input or the output is a vector~\citep{SzaSriPoc16,muandet2017kernel,petersen2019frechet}, and thus does not apply directly to the problem we study here. A complementary line of work extends this setting to \emph{measure-to-measure} (M2M) \emph{regression} by modeling the regression operator as a transport map on the sample space \citep{ghodrati2022distribution,ghodrati2023transportation,Bunne2023NeuralOT,girshfeld2025neural,geuter2022universal,geuter2025ddeqs}. Most of these approaches, however, rely on limited function classes and do not scale effectively to large problems.

\vspace{-0.3em}
\paragraph{Transformers as operators on measures} Transformers provide a substantially more expressive route to modeling nonlinear transformations on the space of probability measures, and have been used successfully to learn functions over sets of point clouds~\citep{lee2019set, yang2019modeling, zhao2021point, haviv2024wasserstein}. A growing body of work analyzes their mean-field limits, emergent dynamics, and approximation properties as operators on spaces of probability measures \citep{sander2022sinkformers,amos2022meta,geshkovski2024emergence,geshkovski2025mathematical,chen2025quantitative,castin2025unifiedperspectivedynamicsdeep}; in particular, under mild conditions, deep transformers are universal approximators of M2M maps \citep{yun2019transformers,alberti2023sumformer,kratsios2021universal,jiang2023approximation,GesRigRui24}.

This literature has remained largely theoretical, and the practical use of transformers as operators between probability measures is still limited. Two recent works explore this direction in specific scenarios: approximating Gaussian mixture models \citep{zimin2025gmm} and learning transformations on cell sets \citep{adduri2025predicting}. Both can be viewed as special cases of our \textit{static} M2M framework, in which a transformer is used as a single-step transport map between measures.

\vspace{-0.3em}
\paragraph{Transformers as dynamic continuous-time transport maps} In contrast to such \textit{static} M2M maps, a recent line of work models these transformations as \textit{dynamic}, continuous-time transport maps whose velocity field is itself distribution-dependent \citep{atanackovic2024meta, haviv2024wasserstein}. The natural geometric object underlying this viewpoint is the \emph{flow map} on the ground space: given a (possibly measure-dependent) time-varying velocity field $v_t$ on $\R^d$, the associated flow map $\Phi_t : \R^d \to \R^d$ sends an initial state $x_0$ to the solution at time $t$ of the ODE $\dot{x}_t = v_t(x_t)$. Flow maps are appealing because they lift dynamics on the ground space to dynamics on the space of measures via pushforward, $\mu_t = (\Phi_t)_\# \mu_0$, thereby reducing the problem of learning an operator on $\mathcal{P}(\R^d)$ to that of learning a vector field on $\R^d$. 

Crucially, this computational reduction comes at no expressiveness cost: under mild conditions, every M2M map can be realized as a continuous-time, measure-dependent flow map, making this class maximally expressive within $\mathcal{P}(\R^d) \to \mathcal{P}(\R^d)$ maps \citep{lavenant2026continuous}. Frameworks such as flow matching \citep{liu2022flow, lipman2022flow, tong2023improving, albergo2023stochastic} then provide a scalable route to training such models in practice. Self-attention has been routinely used to parametrize flow models for point-wise generation trained with flow matching \citep{jing2024generative,ma2024sit,hu2024latent}, but such methods do not address the broader M2M regression problem. We close this gap by introducing a practical and scalable framework for M2M regression through both \textit{static} and \textit{dynamic} transformers.

\vspace{-0.3em}
\paragraph{Contributions} Building on these largely independent lines of research, we formalize the problem of learning nonlinear M2M operators and introduce two distinct transformer-based paradigms, addressing the \textit{static} and \textit{dynamic} settings respectively. Our core contributions are as follows:
\vspace{-0.3em}
\begin{itemize}[noitemsep,nosep,leftmargin=*]
    \item We formalize the nonlinear M2M regression problem through the lens of operators on $\mathcal{P}(\mathbb{R}^d)$, defining the framework via both direct mappings and continuous-time dynamics.
    \item We introduce two practical and scalable approaches to learn such operators: a \textit{static} method that uses transformers to directly approximate the M2M map, and a \textit{dynamic} method, \textit{Transformer Flow Matching} (\tfm), which elegantly connects the measure-dependence of transformers with continuous-time flow matching.
    \item We demonstrate strong empirical performance across several settings, including a large-scale real-world application: predicting colorectal cancer treatment responses in novel patients.
\end{itemize}

\vspace{-1em} 
\section{Measure-to-measure regression}
\vspace{-0.5em} 

In M2M regression, an unknown regression operator maps input measures to output measures. 
The fundamental statistical objects are probability measures rather than individual observations. 
The dataset consists of $n$ pairs, where for each index $i=1,\dots,n$, we observe two finite point clouds on \smash{$\R^d$, $\cX_i = \{x_{i1},\dots,x_{iN}\}$,  $\cY_i = \{y_{i1},\dots,y_{i N}\}$}
which we represent by their empirical measure
\[\mu_i := \frac{1}{N}\sum_{j=1}^{N}\delta_{x_{ij}} \quad \text{and} \quad \nu_i := \frac{1}{N}\sum_{j=1}^{N}\delta_{y_{ij}}\]
Each observation is thus a pair \smash{$(\mu_i,\nu_i)\in\mathcal{P}(\R^d)\times\mathcal{P}(\R^d)$}, and the dataset consists of $n$ such pairs.\footnote{We assume for simplicity that both point clouds have the same number of elements. Extending our setup to point clouds of different cardinality is straightforward as the points $x_{ij}$ and $y_{ij}$ do not need to be paired.} %

The measures $\mu_i$ and $\nu_i$ are treated as the primary statistical objects---elements of $\mathcal{P}(\R^d)$ with finite support---rather than as finite-sample approximations of latent population-level distributions. We regard $(\mu_i,\nu_i)$ as independent copies of a random pair $(\mu,\nu)\in\mathcal{P}(\R^d)\times\mathcal{P}(\R^d)$ satisfying the structural relation $\nu = \cT^\star(\mu)$
for some measurable \emph{operator} $\cT^\star : \mathcal{P}(\R^d)\to\mathcal{P}(\R^d)$. In our motivating application, $\mu$ and $\nu$ correspond to pre- and post-intervention distributions, and $\cT^\star$ represents the effect of a fixed intervention at the distributional level. 

The goal of \emph{measure-to-measure regression} is to estimate $\cT^\star$ from the observed samples \smash{$\{(\mu_i,\nu_i)\}_{i=1}^n$}.
Although the observed measures are finitely supported, $\cT^\star$ is defined on the full space $\mathcal{P}(\R^d)$. Consequently, M2M regression is an extrapolation problem on the space of measures: a good estimator should generalize beyond the empirical supports seen at training time, including to measures with different support sizes or to smooth limits thereof.
The main challenge in this problem is to develop a function class for M2M maps that is simultaneously expressive, computationally scalable, and easy to implement.

\section{Transformers for Measure-to-measure Regression}

In this section, we outline the connections between transformers and M2M regression.  We begin with a brief overview of transformer primitives needed to place these constructions in a common framework (\cref{sec:transformers}), then we discuss M2M operators (\cref{sec:m2m_operators}), and lastly we introduce two novel methods which exploit the measure dependence of transformers to solve M2M regression tasks (\cref{sec:static_m2m} and \cref{sec:dynamic_m2m}).

\subsection{Transformers}\label{sec:transformers}A transformer~\citep{vaswani2017attention} operates on a finite collection of tokens 
\smash{$x_1,\dots,x_N \in \R^d$} by composing $T$ blocks, each combining self-attention, a pointwise feedforward map (MLP), normalization, and residual connections.
For a single attention head, given matrices \smash{$Q,K,V$}, the attention output for token $x_i$ is \[\mathrm{Attn}(x_i;\{x_1, \ldots, x_N\}) = V\sum_{j=1}^N x_j\frac{\exp(\langle Q x_i, K x_j\rangle)}{ \sum_{l=1}^N \exp(\langle Q x_i, K x_l\rangle)}\]
Because attention does not depend on the order of the tokens $x_1,\dots,x_N$, we can equivalently write \smash{$\mathrm{Attn}(x_i;\mu_N)$} that conditions on the empirical measure,\footnote{This \emph{mean-field} structure is characteristic of \emph{encoder} architectures. The causal masking used in \emph{decoder} architectures for large language models breaks this symmetry and is not considered here.} 
\smash{$\mu_N := \frac{1}{N}\sum_{j=1}^N \delta_{x_j}$}.
More generally, for any measure $\mu$ on $\R^d$ and any \smash{$x\in\R^d$} we have, 
\begin{equation}
\label{eq:attcont}
  \mathrm{Attn}(x;\mu) :=
\frac{V\int y \exp(\langle Q x, K y\rangle)\, d\mu(y)}
{\int \exp(\langle Q x, K y\rangle)\, d\mu(y)}.  
\end{equation}
This view of attention emphasizes why transformers are particularly well-suited to M2M regression. Including the MLP, normalization, residual connections, and stacking blocks yields a discrete-time interacting particle system whose dynamics depend on the empirical distribution of the tokens. 
Abstractly, a transformer can thus be viewed as a learned, measure-dependent map $F_{\mu_N}(\cdot) : \R^d \to \R^d$ parameterized by the empirical measure $\mu_N$ of the input tokens. 
This perspective has proved fruitful for the mathematical analysis of transformers, beginning with~\citep{sander2022sinkformers} and further developed in
\citep{geshkovski2024emergence,GesRigRui24,geshkovski2025mathematical,castin2025unifiedperspectivedynamicsdeep,CasAblPey24,chen2025critical,chen2025quantitative,chen2025clustering}. In practice, $F_{\mu_N}$ takes as input a set of samples (tokens) \smash{$x_1,\dots,x_N \in \R^d$} which defines the empirical measure $\mu_N$ and induces measure-dependence by acting on all tokens as a collective via the self-attention mechanism.

\subsection{Measure-to-measure operators} 
\label{sec_transfomers_for_m2m} \label{sec:m2m_operators} 
M2M operators admit a variety of structural forms. A classical choice is a \emph{Markov (linear) operator} $\cT : \mathcal P(\R^d) \to \mathcal P(\R^d)$ of the form $\cT(\mu)(A) = \int K(x,A)\, d\mu(x)$, induced by a Markov kernel $K$. A more recent line of work~\citep{Bunne2023NeuralOT} parametrizes $\cT$ instead as a pushforward $\cT(\mu) = f_\# \mu$ for some map \smash{$f:\R^d \to \R^d$}, with the canonical choice being an optimal transport (OT) map; see e.g.~\citet{CheNilRig25}. Neither family escapes a fundamental expressivity limitation, however, even when the two are combined: whenever two input measures have overlapping supports, so must their images under any Markov operator or pushforward map \citep[Section~1.2]{GesRigRui24}.

In this work, we depart from the linear setting and consider nonlinear M2M operators. Specifically, we consider a parametric class of pushforward operators where the pushforward map can explicitly depend on the source measure itself. Such operators take the form
\begin{equation}
    \label{eq:push}
    \cT(\mu) = (f_\mu)_\# \mu .
\end{equation}
Notably, the existence and regularity of such transport maps were recently formalized by \cite{lavenant2026continuous}, who showed  that Lipschitz continuous operators $\cT$ can always be represented as~\eqref{eq:push} with a continuous pushforward map $f_\mu$---omitting trivial obstructions associated to mass splitting.

We consider two classes of approaches for learning the nonlinear M2M operator in \cref{eq:push}: modeling $f_\mu$ directly via a one-step transport map (\textit{static} M2M regression; \cref{sec:static_m2m}) and modeling $f_\mu$ via a vector field (\textit{dynamic} M2M regression; \cref{sec:dynamic_m2m}). 
We implement $f_\mu$ using transformers, which inherently provide the corresponding measure-dependent maps required to approximate the operator in \cref{eq:push}. Sufficiently deep transformers are universal approximators of M2M operators~\citep{yun2019transformers,alberti2023sumformer,kratsios2021universal,chiang2023tighter,jiang2023approximation,edelman2022inductive,jiang2023brief,wang2024understanding,petrov2024prompting,sander2024towards,GesRigRui24}, and as a result provide a strong basis for our approaches.

\subsection{\textit{Static} transformers for M2M regression}\label{sec:static_m2m}We begin with a natural static formulation where we set the pushforward map in~\eqref{eq:push} to $f_\mu = F_\mu$, where $F_\mu$ is estimated using a  transformer, \smash{$F_\mu^\theta(\cdot): \mathbb{R}^d \rightarrow \mathbb{R}^d$}. We use a standard transformer architecture detailed in \cref{app:arch}. This induces an M2M operator $\cT$ which maps the empirical measure of the inputs to that of the outputs in one step, i.e. 
\[
\cT: \mu_N=\frac{1}{N}\sum_{i=1}^N \delta_{x_i}\;\longmapsto\; (F_{\mu_{\scriptscriptstyle N}}^\theta)_\# \mu_N =\frac{1}{N}\sum_{i=1}^N \delta_{F_{\mu_{ N}}^\theta (x_i)}
\] 
This proposal to use a deep transformer is motivated by their measure-dependent characteristics established in \cref{sec:transformers}. In the following, we introduce a practical and efficient algorithm for training such models.

\paragraph{\textit{Static} M2M training objective} Recall that we have a dataset of measure pairs $\{(\mu_i, \nu_i)\}_{i=1}^n$. By sampling an index $i$ uniformly at random from $\{1, \dots, n\}$ and subsequently drawing mini-batches $\Tilde{\mu}_i \sim \mu_i$ and $\Tilde{\nu}_i \sim \nu_i$, we use stochastic gradient descent to minimize the empirical loss
\begin{equation}
    \mathcal{L}_{\text{M2M}}(\theta) = \frac1n \sum_{i=1}^n \mathbb{E}_{\Tilde{\mu}_i \sim \mu_i, \Tilde{\nu}_i \sim \nu_i} \left[\mathbf{D}\left( \left(F^\theta_{\Tilde{\mu}_i}\right)_\# \Tilde{\mu}_i , \Tilde{\nu}_i\right) \right] .
    \label{eq:loss_static}
\end{equation}
In practice, any differentiable distributional loss $\mathbf{D}$ can be used. In this work, we consider a variety of distributional losses, including the maximum mean discrepancy (MMD) distance, energy distance (ED), and the  Wasserstein distances $\mathcal{W}_1, \mathcal{W}_2$.  
We use the Geometric Loss Functions package \citep{feydy2019interpolating} to compute differentiable distributional losses, and approximate $\mathcal{W}_1$ and $\mathcal{W}_2$ via the Sinkhorn algorithm \citep{cuturi2013sinkhorn}. 
This mirrors \citet{genevay2018learning}, who minimize Sinkhorn divergences
to learn a single map from fixed noise to data. In contrast, our static approaches learn measure-dependent pushforward maps amortized across many measure pairs.
We label these methods \texttt{M2M-MMD}, \edmap, \texttt{M2M-$\mathcal{W}_1$}, and \texttt{M2M-$\mathcal{W}_2$}, respectively. We include detailed definitions of the respective distributional distances in \cref{ap:metrics_and_losses}. 
We minimize \eqref{eq:loss_static} using the Adam optimizer.

As we demonstrate empirically below, this static approach is already competitive across a wide range of M2M tasks. It nevertheless inherits a well-known obstacle: estimating and optimizing distances between probability measures is a notoriously hard statistical problem, especially in high dimensions \citep{si2020quantifying, gao2023finite}. Indeed, this is precisely the difficulty that score matching was developed to circumvent in the generative-modeling literature, by replacing the optimization of an intractable distributional objective with a tractable pointwise regression on the score function. Motivated by the same principle, we introduce a second, \emph{dynamic} approach in which time-dependent transformers model a velocity field rather than an end-to-end pushforward map and are trained via a pointwise regression objective, thereby sidestepping distributional losses entirely.

\subsection{\textit{Dynamic} transformers for M2M regression}\label{sec:dynamic_m2m}In this section, we introduce a novel approach that uses a time-dependent transformer and conditional flow matching \citep{lipman2022flow, tong2023improving} to model the vector field for the iterative evolution of $\mu_i$ to $\nu_i$. The resulting model is easier to optimize and more expressive, leveraging auto-regressive inference---iterating for multiple steps with the transformer to produce $\nu_i$---in contrast to the static map that takes one step to output $\nu_i$. 

\paragraph{The \textit{dynamic} formulation} We adopt a dynamical viewpoint inspired by Otto calculus and the geometry of the Wasserstein space; see, e.g., \citet{CheNilRig25}. Recall that the $2$-Wasserstein space $\mathcal P_2(\R^d)$ admits a formal differential structure in which tangent vectors at a measure $\mu$ are identified with vector fields $\chi:\R^d\to\R^d$ satisfying suitable integrability conditions. With this structure, a curve $(\mu(t))_{t\in[0,T]}$ is absolutely continuous in the $2$-Wasserstein metric if and only if  for a time-dependent velocity field $\chi_t[\mu(t)] \in L^2(\mu(t))$, it satisfies a measure-dependent continuity equation of the form
\begin{equation}\label{eq:cont}
    \partial_t \mu(t) + \nabla\!\cdot\!\bigl(\mu(t)\,\chi_t[\mu(t)]\bigr)=0 .
\end{equation}
In classical OT, admissible velocity fields are further restricted to gradients of scalar potentials, yielding a Riemannian structure on $\mathcal P_2(\R^d)$. We do not impose this restriction and allow for general, possibly non-gradient vector fields, which still generate absolutely continuous curves while significantly enlarging the class of admissible dynamics.

Given a family of velocity fields $(\chi_t[\mu(t)])_{t}$ and an initial condition $\mu(0)$, the continuity equation uniquely determines a curve $(\mu(t))_{t\in[0,T]}$. The corresponding M2M operator is defined as the \emph{flow map} \smash{$\cT:\quad \mu(0) \longmapsto \mu(T)$}.
As established in Section~\ref{sec:transformers}, transformers naturally define measure-dependent maps from $\R^d$ to $\R^d$ and therefore have precisely the structure of admissible velocity fields $\chi_t[\mu(t)]$. This motivates our key proposal: to \emph{use a time-dependent transformer to parameterize the velocity fields} in the continuity equation. 
Integrating the resulting dynamics yields an absolutely continuous curve in $\mathcal P(\R^d)$ and, through its endpoint, a nonlinear M2M operator.

\textbf{\textit{Dynamic} M2M training objective.} 
To learn $\chi_t[\mu(t)]$, we introduce \textit{Transformer Flow Matching} (\tfm). 
As established in Section~\ref{sec:transformers}, transformers naturally process empirical measures as sets of tokens, allowing them to define measure-dependent maps without requiring explicit conditioning networks. This makes them perfectly suited to parameterize the velocity fields in the measure-dependent continuity equation, which we denote as a time-dependent transformer $\chi_t^\theta[\mu(t)](\cdot) : [0, 1] \times \mathbb{R}^d \rightarrow \mathbb{R}^d$. 
To train $\chi_t^\theta[\mu(t)]$, we extend Conditional Flow Matching (\texttt{CFM}) \citep{lipman2022flow, tong2023improving} to marginalize over the training set of measure pairs. This is akin to the setup of \cite{atanackovic2024meta}, but eschewing the need for a secondary model to capture distribution-dependence.
Specifically, let $\gamma_i \in \Pi(\mu_i, \nu_i)$ denote a chosen coupling between the $i$-th source and target distributions. For a mini-batch-sampled pair $(x,y) \sim \gamma_i$, we define a continuous-time interpolation path $\phi_t(x, y)$, with $\phi_{t=0}(x, y) = x$ and $\phi_{t=1}(x, y) = y$, inducing the batched measure-level path $\Tilde{\mu}_{i,t} = (\phi_t)_\# \gamma_i$. The parameters $\theta$ are then optimized by minimizing the following empirical loss:
\begin{equation} \label{eq:loss_tfm}
    \vspace{-0.1em}
    \mathcal{L}_{\mathrm{TFM}}(\theta) = \frac1n \sum_{i=1}^n\mathbb{E}_{(x,y) \sim \gamma_i}\int_0^1 \bigl\| \chi_t^\theta[\Tilde{\mu}_{i,t}](\phi_t(x, y)) - \frac{\partial}{\partial t}\phi_t(x, y) \bigr\|^2 dt.
    \vspace{-0.1em}
\end{equation}
\begin{wrapfigure}{r}{0.5\textwidth}
  \begin{minipage}{\linewidth} 
    \vspace{-1.2em} %
    
    \begin{algorithm}[H] 
      \caption{(\textit{Training step})}
      \label{alg:tfm_training}
      \begin{lstlisting}[belowskip=-0.2em, aboveskip=-0.05em, mathescape=true, basicstyle=\footnotesize\linespread{0.4}\selectfont\ttfamily]
# Inputs:
#  Dataset: $\color{pastel_green}\mathcal{M} = \{(\mu_i, \nu_i)\}_{i=1}^n$ 
#       with $\color{pastel_green}\mu_i \color{pastel_green}\in \mathbb{R}^{\color{pastel_green}N_{\mu_i} \times \color{pastel_green}d}$, $\color{pastel_green}\nu_i \color{pastel_green}\in \mathbb{R}^{\color{pastel_green}N_{\nu_i} \times \color{pastel_green}d}$
#  $\color{pastel_green}b$:(num measures) $\color{pastel_green}m$:(particles per measure) 
#  $\color{pastel_green}\theta$: Transformer parameters, $\color{pastel_green}\chi_t^\theta[\mu(t)](\cdot)$

x, y = sample_measures($\mathcal{M}$, $b$, $m$) 
# tuple ([$\color{pastel_green}b, \color{pastel_green}m, \color{pastel_green}d$], [$\color{pastel_green}b, \color{pastel_green}m, \color{pastel_green}d$])  

x, y = ot_couplings(x, y)  
# if using mini-batch OT 

t = sample_uniform(0, 1) # sample $\color{pastel_green}b$ $\color{pastel_green}t$'s
z = t * y + (1-t) * x 

v_target = y - x
v_pred = Transformer(z, t, $\theta$)

loss = MSE(v_target, v_pred)
$\theta$ = update(loss, $\theta$)
      \end{lstlisting}
    \end{algorithm}

    \vspace{-1.5em} %

    \begin{algorithm}[H]
      \caption{(\textit{Inference / sampling})}
      \label{alg:tfm_inference}
      \begin{lstlisting}[belowskip=-0.2em, aboveskip=-0.05em, mathescape=true,basicstyle=\footnotesize\linespread{0.4}\selectfont\ttfamily]
# Inputs:
#   Test Measure: $\color{pastel_green} \mu_j \color{pastel_green}\in \mathbb{R}^{\color{pastel_green}N_{\mu_j} \times \color{pastel_green}d}$ , 
#   $\color{pastel_green}\theta$: Transformer parameters, $\color{pastel_green}\chi_t^\theta[\mu(t)](\cdot)$
#   num_steps: number of simulation steps 

dt = 1/num_steps, t = 0
z = $\mu_j$ # [$\color{pastel_green}1, \color{pastel_green}N_{\mu_j}, \color{pastel_green}d$]

for _ in range(num_steps):
    z += Transformer(z, t, $\theta$) * dt
    t += dt

return z # approximates true $\color{pastel_green}\nu_j$ 
      \end{lstlisting}
    \end{algorithm}
    
    \vspace{-5em} %
  \end{minipage}
\end{wrapfigure}
Since $\chi_t^\theta$ takes $\tilde{\mu}_{i,t}$ as input, the measure acts
as the condition, so standard guarantees for conditional generation
\citep{dao2023flow, zheng2023guided, isobe2024extended} apply to
\eqref{eq:loss_tfm}. In particular, its minimizer generates the path
$\tilde{\mu}_{i,t}$ for every pair, transporting $\mu_i$ at $t=0$ to $\nu_i$
at $t=1$.

In this work, we focus on linear paths of the form \smash{$\phi_t(x, y) = (1 - t)x + ty$}, which yields the constant velocity \smash{$\frac{\partial}{\partial t}\phi_t(x, y) = y - x$}. 
This induces the corresponding measure-level path \smash{$\Tilde{\mu}_{i,t} = ((1-t)\phi_{t=0} + t \phi_{t=1})_\sharp \gamma_i$}. 
We restrict the couplings $\gamma_i$ to OT plans, which can be efficiently approximated via mini-batch OT \citep{nguyen2021transportation, fatras2021minibatch}. We optimize \cref{eq:loss_tfm} using the Adam optimizer and provide the pseudo-code for our training algorithm in \cref{alg:tfm_training}. Extensions to other conditional paths are left for future work.

At test-time, given an \textit{unseen} source measure $\mu_j$, we evaluate \tfm by predicting the corresponding $\nu_j$ by solving the empirical ODE \smash{$\frac{d}{dt} z_t = \chi_t^\theta[\mu_t](z_t)$} from $t=0$ to $t=1$ (e.g., via Euler integration). This realizes a prediction for the unseen measure \smash{$\hat{\nu}_j = (f_{\mu_j})_\sharp \mu_j$}. We provide pseudo-code for inference/sampling in \cref{alg:tfm_inference}. 

We provide pseudo-code for training and inference of the \textit{static} models in \cref{ap:metrics_and_losses}. We additionally consider an \texttt{M2M-OTMSE} variant of the static approach that pairs source and target samples through a mini-batch OT coupling and minimizes
$\mathbb{E}_{(x,y) \sim \gamma_i} \| F_{\Tilde{\mu}_i}^\theta(x) - y \|^2$.
See \cref{app:arch_and_hparams} for details regarding architecture and hyperparameters for \textit{static} and \textit{dynamic} models.

\vspace{-0.5em}
\section{Related Work}
\label{sec:related}
\vspace{-0.5em}

Our work touches on three lines of research that have largely developed independently: 
(i) M2M regression and learning M2M operators (discussed in \cref{sec:intro}), 
(ii) transformers as nonlinear operators on measures,
and (iii) extensions of flow matching to the space of probability measures. 
Here we review the most closely related works and position our contribution relative to them.

\paragraph{Transformers as nonlinear operators on measures}
Transformers have successfully been applied to learning complex functions on point cloud data~\citep{lee2019set, yang2019modeling, zhao2021point, haviv2024wasserstein}.
There is a growing body of work exploring transformers as nonlinear, permutation-equivariant operators acting on empirical measures \citep{GesRigRui24, geshkovski2025mathematical, lavenant2026continuous}. 
However, there is less work on practically using transformers as learned operators between probability measures. \citet{zimin2025gmm} used a transformer for M2M regression for Gaussian mixture distributions, focused around the clustering capabilities of transformers optimized with a squared quadratic Wasserstein distance. \citet{adduri2025predicting} developed a more complex procedure specifically for predicting the transformations of cell sets by optimizing a transformer via an ED loss. The approach differs, in that in our own experiments with cell data, we directly input the cell description into the transformer, bypassing the cell (state) embedding model.

\paragraph{Learning flows on spaces of probability measures} 
A related line of work extends flow matching to the space of probability measures, and naturally splits according to the problem it addresses: \emph{M2M regression} or \emph{distribution alignment}. We discuss each in turn.

For M2M regression---the problem we consider---the most closely related approach is \emph{Meta Flow Matching} (\texttt{Meta-FM}) of \citet{atanackovic2024meta}, which jointly learns a distribution-dependent vector field and a distribution embedding via flow matching. This setup requires alternating optimization between two separate networks (the vector field and the embedding model), and the distributional dependence enters only through the initial condition at $t=0$ rather than along the entire trajectory. Concurrent work by \citet{fishman2026distribution} pursues a closely related direction, using distributional encoders to model distribution-conditioned transport in the spirit of \texttt{Meta-FM}.

A second, distinct body of work addresses \emph{distribution alignment} on $\mathcal{P}(\R^d)$, which is best understood by analogy with the Euclidean case. On $\R^d$, generative modeling learns to sample from an unknown distribution given i.i.d.\ samples, whereas regression learns a specific map $f:\R^d \to \R^d$ from input--output pairs $(x_i, y_i)$. Distribution alignment is the natural lift of \emph{generative modeling} to $\mathcal{P}(\R^d)$: the training data are themselves measures $\mu_1,\dots,\mu_n$, viewed as i.i.d.\ samples from an unknown \emph{meta-distribution} on $\mathcal{P}(\R^d)$, and the goal is to sample new measures from this same meta-distribution. M2M regression, by contrast, is the lift of \emph{regression}: the data come as input--output pairs of measures $(\mu_i, \nu_i)$, and the goal is to recover the operator that maps each $\mu_i$ to its specific $\nu_i$. In this distribution-alignment category, \citet{haviv2024wasserstein} introduce \emph{Wasserstein Flow Matching} (\texttt{WFM}), which lifts the flow matching framework of \citet{lipman2022flow} from $\R^d$ to $\mathcal{P}(\R^d)$ using a permutation-equivariant velocity field, typically a transformer. \citet{amos2022meta} pursue a similar goal via meta--optimal transport using input-convex neural networks, and \citet{sakalyan2025modeling} apply variational flow matching with transformers to the generation of complex biological states. Because these methods are designed to match meta-distributions rather than individual input--output pairs, they yield sub-optimal solutions when applied directly to M2M regression; a gap we document empirically in \cref{sec:objects}. A related direction uses similar tools to learn distributional \emph{representations} \citep{fishman2025generative}.

\vspace{-1em}
\section{Experiments}
\label{sec:expts_main}

\vspace{-0.5em}

In this section, we evaluate our method's ability to generalize to previously \textit{unseen} test measures. For this we consider three experimental settings: \textbf{(i: \textit{synthetic multi-measure objects}}, \cref{sec:objects}) a dataset of $M$ measure pairs, (\textbf{ii: \textit{dynamics}}, \cref{sec:exp_mkvs}) simulated population dynamics on multi-measure-multi-timepoint McKean-Vlasov systems, and (\textbf{iii: \textit{a real world application}}, \cref{sec:patients}) prediction of response to cancer treatment for \textit{unseen} patients. In all settings, the objective is to evaluate generalization performance on unseen measures.
We describe our individual experimental setups, implementation details, and findings in the following sections. 

\textbf{Baselines.} We compare our approaches (\tfm, \edmap, \texttt{M2M}-$\mathcal{W}_1$) to both non-measure-dependent and measure-dependent methods. We consider Conditional Flow Matching (\texttt{CFM}) \citep{lipman2022flow, tong2023improving} as the non-measure-dependent baseline.\footnote{We remark that we use ``\texttt{CFM}'' to denote two different variants used separately in each experimental setting. In \cref{sec:objects} and \cref{sec:exp_mkvs} we use mini-batch \texttt{OT-CFM} directly and in \cref{sec:patients} we use \texttt{OT-CFM} with precomputed couplings.}
To deploy \texttt{CFM} in the M2M setting, we equivalently marginalize its loss over the training measure pairs and use the same couplings as \tfm, with all other experimental conditions matched. Unlike \tfm, however, \texttt{CFM} parameterizes the vector field with an MLP that acts on individual points, and therefore cannot capture measure-dependence.

We consider \texttt{Meta-FM} \citep{atanackovic2024meta} as a flow matching-based measure-dependent approach tailored for the problem of M2M regression. We additionally consider \texttt{WFM} as a baseline (which assumes unpaired measures) for the synthetic multi-measure objects experiments (\textbf{i}, \cref{sec:objects}). For the multi-time-point McKean-Vlasov experiments (\textbf{ii}, \cref{sec:exp_mkvs}), we additionally consider two state-of-the-art baselines: the Neural McKean-Vlasov Process (\texttt{NMKV}) \citep{pmlr_nmkv} and the mean-field transformer (\texttt{MF-Transformer}) \citep{mean_field_biswal}. We use the best performing variant of \texttt{NMKV}, the \textit{empirical measure architecture}. We provide further details in \cref{app:baselines}.

\begin{figure*}[t]
    \vspace{-2em}
    \centering
    \begin{subfigure}[b]{0.46\textwidth}
        \centering
        \includegraphics[width=\linewidth]{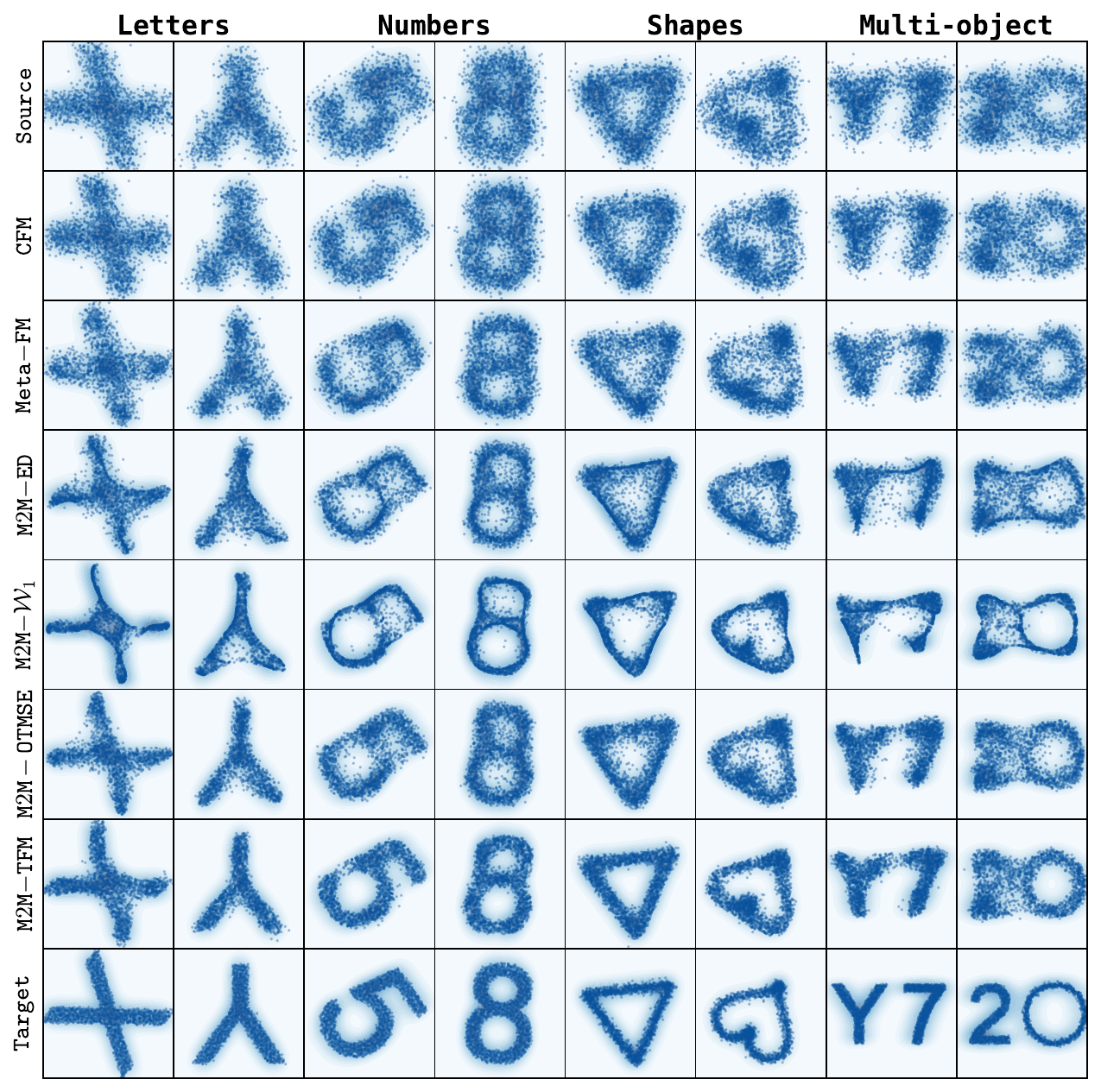}
        \caption{\textbf{Diffusion process}}
        \label{fig:diffusion_plot}
    \end{subfigure}
    \hspace{0.05\textwidth}
    \begin{subfigure}[b]{0.46\textwidth}
        \centering
        \includegraphics[width=\linewidth]{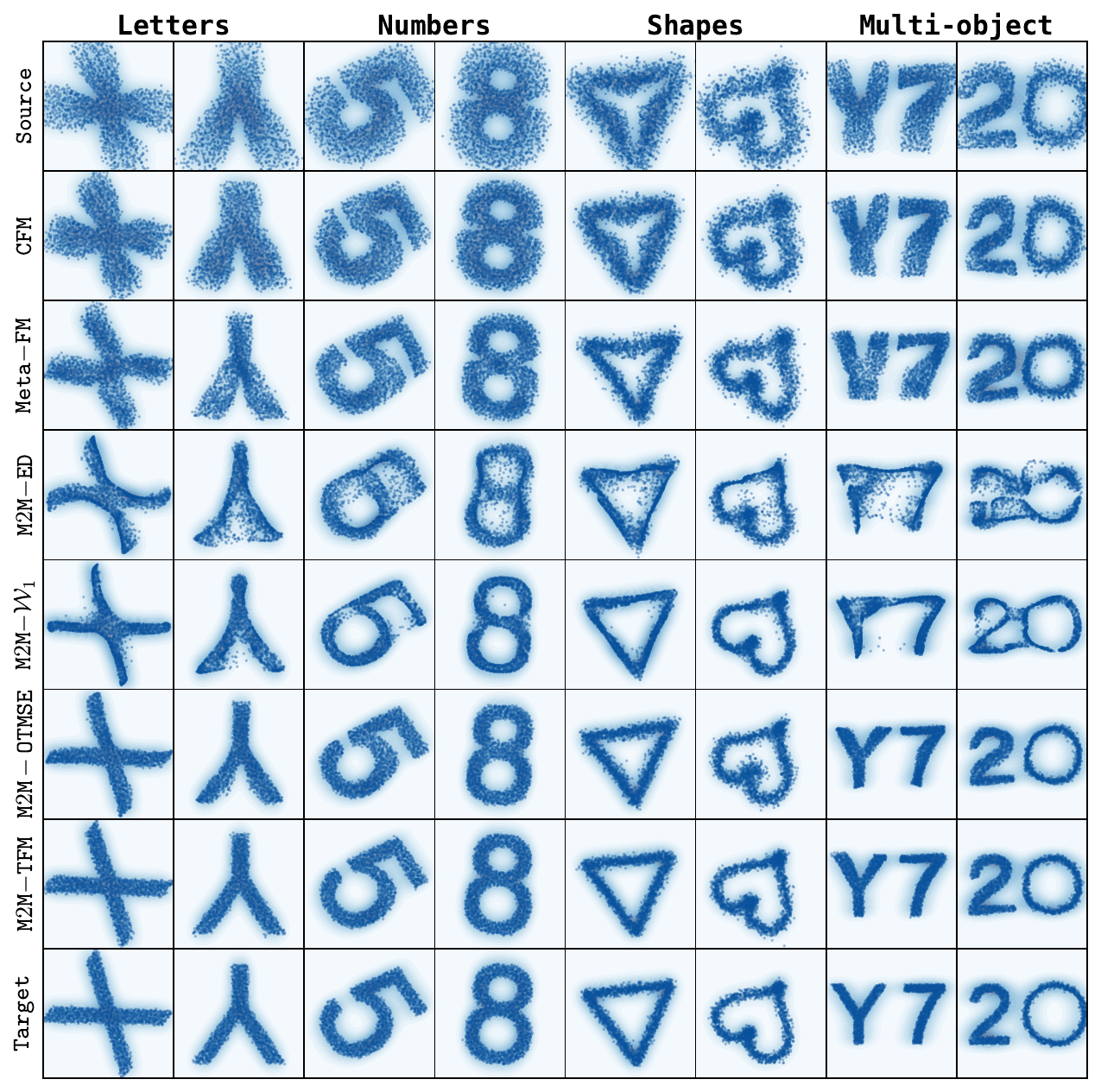}
        \caption{\textbf{Kernel interaction process}}
        \label{fig:kernel_plot}
    \end{subfigure}

    \vspace{-0.3em}
    \caption{\textbf{Visualization of model predictions on multi-measure objects for measures outside of the training set.} Methods are tasked with transporting particles from \textit{source} measures (top row) to \textit{target} measures (bottom row). We show visuals for models trained to reverse (a) a diffusion process, and (b) a kernel interaction process.}
    \label{fig:letters_viz}
    \vspace{-1em}
\end{figure*}

\vspace{-0.5em}
\subsection{Paired multi-measure objects}\label{sec:objects}
\vspace{-0.5em}

\textbf{Setup.} We consider the synthetic letters \textit{silhouettes} setup from \citet{atanackovic2024meta}, with several adjustments to increase the difficulty of the learning problem. 
To construct a dataset of measure pairs \smash{$\mathcal{M} = \{(\mu_i, \nu_i)\}_{i=1}^M$} (with $M=240$), we first define a set of \textit{target} measures $\{\nu_i\}_{i=1}^M$, using 2D letter shapes. 
We use all letters, except ``X'' and ``Y'', with each letter in 10 random orientations. We place $N_i$ particles (approximately $2000 \pm 500$ per-measure) on the silhouettes of each target measure $\nu_i$, and then apply a corruption process to acquire $\mu_i$. 
We use both the forward diffusion from \citet{atanackovic2024meta} and a \textit{kernel interaction} corruption process defined in \cref{ap:letters_expt_details}. Unlike the setup in \citet{atanackovic2024meta}, we shuffle all the samples/particles between the source and target measures of each object silhouette.

Models are tasked with learning the reverse corruption process ($\mu \rightarrow \nu$) and are evaluated on unseen measures. 
Since particles within measures are unpaired, we use mini-batch OT to compute couplings between particles every training step. We evaluate performance on four categories of left-out measures, each with $1500$ to $2000$ particles; \textit{letters} ``X'' and ``Y'', \textit{numbers}, \textit{shapes}, and \textit{multi-object} compositions. We use the $\mathcal{W}_1$ and $\text{ED}$ distributional metrics to quantify performance.

\begin{figure}[t]
  \begin{minipage}[t]{\dimexpr\textwidth-2.15in-1em\relax}
    \vspace{10pt}
    \centering
    \captionof{table}{\textbf{Prediction performance comparison on multi-measure objects.} Mean $\pm$ Std computed over 5 random model seeds and 8 ``\textit{object}'' silhouettes. \textbf{Bold} indicates the best performing method.}
    \label{tab:letters_avg}
    \setlength{\tabcolsep}{3pt}
    \resizebox{\linewidth}{!}{%
    \begin{tabular}{lcccc}
        \toprule
        & \multicolumn{2}{c}{\textbf{\texttt{Diffusion}}} & \multicolumn{2}{c}{\textbf{\texttt{Kernel interactions}}} \\
        \cmidrule(lr){2-3} \cmidrule(lr){4-5} %
        \texttt{\textbf{Method}} & $\mathcal{W}_1$ ($\downarrow$) & ED ($\downarrow$) & $\mathcal{W}_1$ ($\downarrow$) & ED ($\downarrow$) \\
        \midrule
        \multicolumn{5}{l}{\textbf{\textit{Baselines}}} \\
        \texttt{CFM} & $0.2491 \pm 0.0032$ & $0.0071 \pm 0.0004$ & $0.2744 \pm 0.0030$ & $0.0128 \pm 0.0005$ \\
        \texttt{Meta-FM} & $0.2306 \pm 0.0114$ & $0.0082 \pm 0.0017$ & $0.2292 \pm 0.0150$ & $0.0130 \pm 0.0022$ \\
        \texttt{WFM} & $0.5834 \pm 0.1399$ & $0.1201 \pm 0.0813$ & $0.5226 \pm 0.0545$ & $0.0931 \pm 0.0303$ \\
        \midrule
        \multicolumn{5}{l}{\textbf{\textit{Static (ours)}}} \\
        \texttt{M2M-MMD} & $0.2036 \pm 0.0094$ & $0.0073 \pm 0.0012$ & $0.1725 \pm 0.0312$ & $0.0068 \pm 0.0018$ \\
        \edmap & $0.2034 \pm 0.0073$ & $0.0062 \pm 0.0005$ & $0.1847 \pm 0.0283$ & $0.0068 \pm 0.0014$ \\
        \texttt{M2M-}$\mathcal{W}_2$ & $0.2288 \pm 0.0151$ & $0.0102 \pm 0.0023$ & $0.1888 \pm 0.0380$ & $0.0080 \pm 0.0025$ \\
        \texttt{M2M-}$\mathcal{W}_1$ & $0.2092 \pm 0.0055$ & $0.0085 \pm 0.0005$ & $0.1663 \pm 0.0235$ & $0.0069 \pm 0.0009$ \\
        \texttt{M2M-OTMSE} & $0.1680 \pm 0.0032$ & $0.0044 \pm 0.0003$ & $0.1334 \pm 0.0088$ & $0.0053 \pm 0.0009$ \\
        \midrule
        \multicolumn{5}{l}{\textbf{\textit{Dynamic (ours)}}} \\
        \tfm & $\mathbf{0.1300 \pm 0.0009}$ & $\mathbf{0.0023 \pm 0.0001}$ & $\mathbf{0.0780 \pm 0.0020}$ & $\mathbf{0.0020 \pm 0.0002}$ \\
        \bottomrule
    \end{tabular}}
  \end{minipage}\hfill
  \begin{minipage}[t]{2.15in}
    \vspace{9.8pt}
    \centering
    \includegraphics[width=\linewidth]{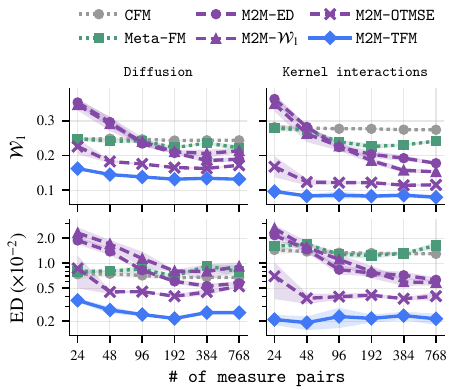}
    \vspace{-12pt}
    \captionsetup{justification=raggedright}
    \captionof{figure}{Ablation over number of training measure pairs on multi-measure objects.} \label{fig:scaling}
  \end{minipage}
  \vspace{-15pt}
\end{figure}

\textbf{Results.} We report quantitative results in \cref{tab:letters_avg} and visualization of model predictions in \cref{fig:letters_viz}. We observe that \tfm outperforms all other approaches across all left-out measure categories, including having lower variance, while our corresponding static approaches outperform baselines. 
We observe that indeed \texttt{WFM} struggles to compete with approaches tailored for the M2M regression problem as it is required to learn the true corresponding \textit{distribution alignment} between measures in addition to the sample-level transport maps. 
We additionally ablate the number of measure pairs used for training (\cref{fig:scaling}), and observe \tfm outperforms all approaches for every number of measure pairs on both systems, performing well with as few as 24 pairs. All models are trained for a fixed number of epochs across all numbers of measure pairs (24 to 768), and we report the mean and standard deviation of $\mathcal{W}_1$ and ED over 3 seeds.
We include extended quantitative results in \cref{tab:letters_extended_diffusion_comparison} and \cref{tab:letters_extended_kernel_interactions_comparison}, extended visualizations in \cref{fig:letters_viz_extended}, and further ablations in \cref{ap:letters_additional_expts}.

\subsection{Multi-timepoint McKean-Vlasov SDEs}\label{sec:exp_mkvs}

We evaluate our approaches for predicting the evolution of synthetic McKean-Vlasov processes \citep{mckean1966class}. We use McKean-Vlasov processes because their dynamics are distribution dependent and therefore a measure-to-measure approach is required for accurate prediction. 

\textbf{Setup.} For each McKean-Vlasov system, models are trained on $70$ measure trajectories, and are evaluated on $10$ held-out trajectories. Each training measure trajectory consists of a $100$ element measure tuple, $(\mu_0, \mu_1, \cdots, \mu_{99})$ representing $100$ time-points in the evolution of the McKean-Vlasov system.  We train the model by regressing between all pairs of time-points independently. At each step we train on a sub-sample $b=16$ of the possible \smash{$\{(\mu_t, \mu_{t+1})\}_{t=0}^{t=98}$} pairs. 
\begin{wrapfigure}{r}{0.6\textwidth}
    \centering
    \vspace{-13pt}
    \begin{tikzpicture}
        \node[inner sep=0pt] (fig) {
            \includegraphics[width=0.6\textwidth]{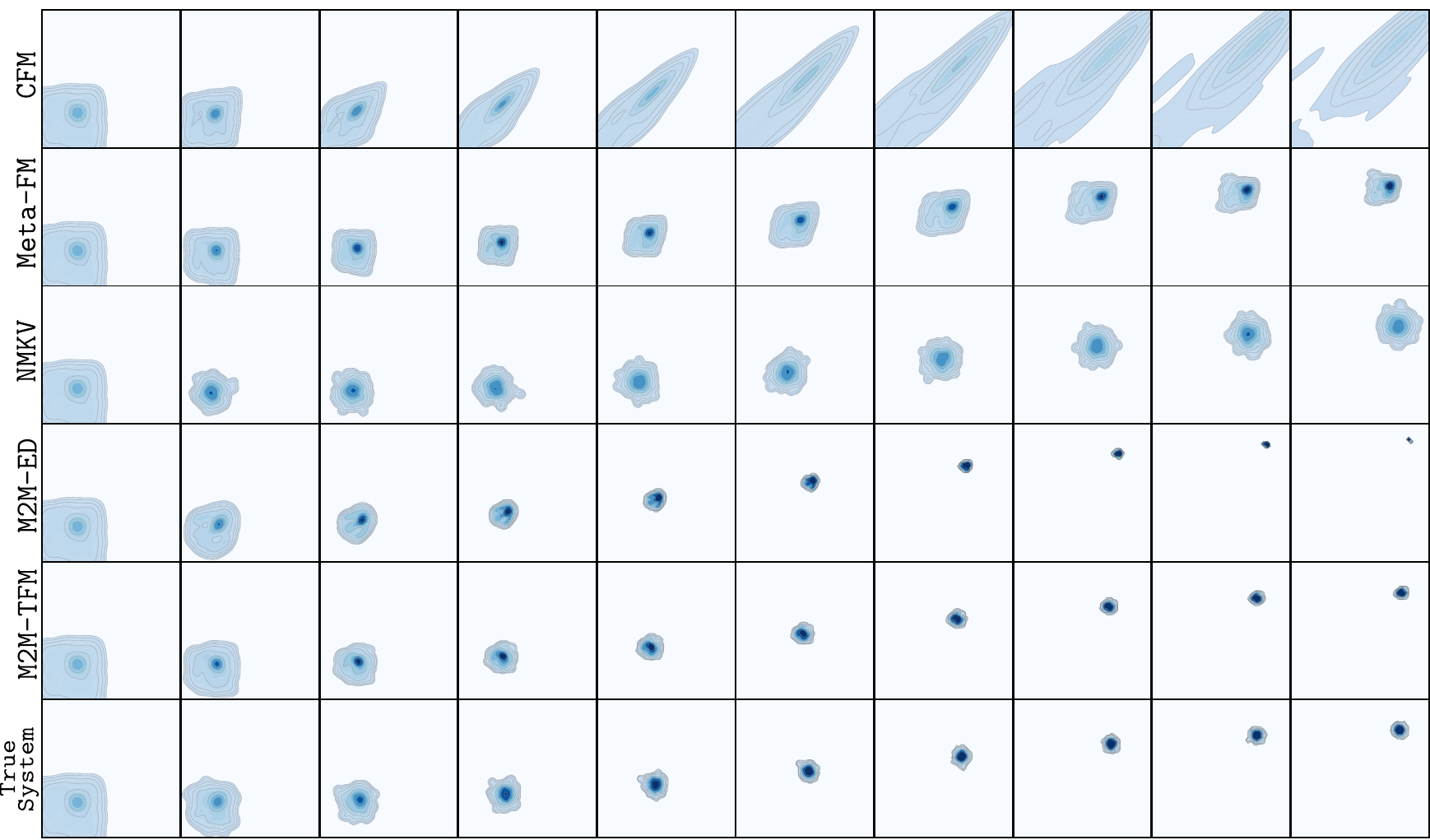}
        };

        \draw[->, thick, >=stealth] 
            ($(fig.north west) + (0, 0.2)$) -- 
            ($(fig.north east) + (0, 0.2)$)
            node[midway, above] {$t$};
            
    \end{tikzpicture}
    \vspace{-1em}
    \caption{\textbf{Visualization of various model predictions on $2$ dimensional Kuramoto system.} We sample new initial conditions then repeatedly apply the models, and the true dynamics. For static methods we only show visualization for \texttt{M2M-ED}.}
    \label{fig:kuramoto_2d_viz}
    \vspace{-1.7em}
\end{wrapfigure}
For evaluation the model receives a new (unseen) measure $\mu_0$ and must predict $\hat{\mu}_0, \hat{\mu}_1, \cdots, \hat{\mu}_{99}$, with evaluation metrics averaged over all time-points except the first. 
We train and evaluate on three modified McKean-Vlasov systems across three dimensional settings: a Kuramoto, Mean-Field Atlas, and FitzHugh-Nagumo system, each in $2$-, $50$-, and $100$-dimensional instantiations. We adapt the McKean-Vlasov processes from \citet{pmlr_nmkv}, with modified dynamics and evaluation in new measures. Details about each of the systems, the initial conditions used, and modifications to the dynamics are provided in \cref{ap:mvsde_expt_details}. Hyperparameters for all methods are in \cref{app:arch_and_hparams}.
\begin{table*}[t]
\centering
\caption{\textbf{Prediction performance comparison on $\mathbf{100}$-dimensional McKean-Vlasov systems.} Mean$\pm$Std computed over $6$ random seeds. \textbf{Bold} indicates the best performing method. Extended results are in \cref{app:additional_mkv}.}
\resizebox{\textwidth}{!}{%
\begin{tabular}{lccccccccc}
\toprule
 & \multicolumn{2}{c}{\textbf{\texttt{Kuramoto}}} & \multicolumn{2}{c}{\textbf{\texttt{FitzHugh-Nagumo}}} & \multicolumn{2}{c}{\textbf{\texttt{Atlas}}} \\
\cmidrule(lr){2-3} \cmidrule(lr){4-5} \cmidrule(lr){6-7}
\texttt{Method} & $\mathcal{W}_1$ ($\downarrow$) & ED ($\downarrow$) & $\mathcal{W}_1$ ($\downarrow$) & ED ($\downarrow$) & $\mathcal{W}_1$ ($\downarrow$) & ED ($\downarrow$) \\
\midrule
\multicolumn{5}{l}{\textbf{\textit{Baselines}}} \\[1mm]
\texttt{CFM} & $11.850 \pm 0.085$ & $7.215 \pm 0.105$ & $14.929 \pm 0.056$ & $0.651 \pm 0.008$ & $20.502 \pm 0.080$ & $6.729 \pm 0.040$ \\
\texttt{MF-Transformer} & $5.462 \pm 0.098$ & $3.315 \pm 0.202$ & $23.435 \pm 0.254$ & $3.305 \pm 0.276$ & $17.175 \pm 0.026$ & $8.721 \pm 0.026$ \\
\texttt{NMKV} & $23.441 \pm 1.559$ & $27.716 \pm 3.117$ & $22.191 \pm 0.109$ & $3.376 \pm 0.097$ & $21.133 \pm 0.077$ & $12.835 \pm 0.131$ \\
\texttt{Meta-FM}& $4.532 \pm 0.199$ & $3.258 \pm 0.320$ & $14.801 \pm 0.175$ & $1.028 \pm 0.053$ & $18.656 \pm 0.093$ & $7.641 \pm 0.042$ \\
\midrule
\multicolumn{5}{l}{\textbf{\textit{Static (ours)}}} \\[1mm]
\texttt{M2M-}$\mathcal{W}_1$ & $5.204 \pm 0.436$ & $7.875 \pm 0.872$ & $14.757 \pm 0.077$ & $7.610 \pm 0.163$ & $20.358 \pm 0.044$ & $23.132 \pm 0.086$ \\
\texttt{M2M-ED} & $14.400 \pm 0.065$ & $17.456 \pm 0.326$ & $22.320 \pm 0.390$ & $3.653 \pm 0.293$ & $32.217 \pm 0.268$ & $12.429 \pm 0.128$ \\
\texttt{M2M-OTMSE}& $6.751 \pm 1.154$ & $10.964 \pm 2.306$ & $17.311 \pm 0.297$ & $11.812 \pm 0.608$ & $21.407 \pm 0.367$ & $25.246 \pm 0.738$ \\
\midrule
\multicolumn{5}{l}{\textbf{\textit{Dynamic (ours)}}} \\[1mm]
\tfm & $\mathbf{3.005 \pm 0.037}$ & $\mathbf{2.299 \pm 0.092}$ & $\mathbf{12.687 \pm 0.084}$ & $\mathbf{0.548 \pm 0.033}$ & $\mathbf{15.940 \pm 0.055}$ & $\mathbf{6.633 \pm 0.130}$ \\
\bottomrule
\end{tabular} 
\label{fig:mkv_results}
}
\vspace{-1.5em}
\end{table*}

\paragraph{Results} We report results in \cref{fig:mkv_results}, on the $100$-dimensional systems; results for all three processes in $2, 50$ and $100$ dimensions are in \cref{ap:additional_expts}. \texttt{M2M-TFM} outperforms in almost all of the systems, reflecting expressivity and ability to scale more consistently with dimension. \texttt{NMKV} performs worse than one might expect, potentially because it is built for generalizing within a measure, rather than between measures. The \texttt{Meta-FM} baseline often exhibits worse performance, potentially due to dependence only on the initial time-point and requiring optimization of two models. We visualize the model predictions of the $2$-dimensional Kuramoto system in \cref{fig:kuramoto_2d_viz}. In \cref{app:additional_mkv}, we provide extended results including; an ablation on the number of simulation steps for \tfm, and mean cosine distances between the true and the learned vector fields for a subset of models.

\vspace{-0.5em}
\subsection{Patient-specific treatment response for colorectal cancer}\label{sec:patients}

\vspace{-0.5em}
\paragraph{Setup} Here, we consider the problem of predicting cellular response in novel patients. 
We use a large-scale multi-donor patient-derived organoid (PDOs)\footnote{A patient-derived organoid (PDO) is a biological model system which is constructed using donor cells from a patient. PDOs capture both the physical/functional characteristics and the genetic profile of the original tissue.} dataset introduced by \citet{zapatero2023trellis}. 
This dataset comprises single-cell (the \textit{samples}) mass-cytometry readouts over 44 bio-markers (features) from PDOs derived from 10 patients. Response screens are conducted over varying treatments, tumor micro-environment conditions, and replicated experiments. 
As a result, the dataset consists of a large quantity of treatment response populations coupled with a corresponding control population (the \textit{measure} pairs $(\mu_i, \nu_i)$). 
We follow the pre-processing steps from \citet{atanackovic2024meta} and define three evaluation splits based on held-out patients (labeled PDO-21, PDO-27, and PDO-75). We use all 44 bio-markers available.
After applying additional quality control steps (detailed in \cref{ap:pdo_expt_details}), the resulting quantity of train/test measure pairs are ($\text{PDO-21:} 393/62,\text{PDO-27:} 404/51, \text{PDO-75:} 390/65$). 
\begin{wraptable}{r}{0.58\textwidth}
  \vspace{-0.9em} %
  \caption{\textbf{Comparison of method prediction performance on \textit{unseen} patients on the patient-derived organoid dataset.} We report mean and standard deviation across 3 left-out patients and 4 model seeds per-patient. \textbf{Bold} indicates the top performer and \underline{underline} indicates the second best performer.}
  \label{tab:pdo_table}
  \vspace{-1em}
  \begin{center}
    \footnotesize %
    \scshape
    \resizebox{\linewidth}{!}{%
      \begin{tabular}{lccc}
        \toprule
         & $\mathcal{W}_1 \downarrow$ & $\text{ED} \downarrow$ & $r^2 \uparrow$ \\
        \midrule
        \multicolumn{4}{l}{\textbf{\textit{Baselines}}} \\[1mm]
        \texttt{CFM} & $4.4099 \pm 0.2580$ & $\underline{0.4828} \pm \underline{0.1695}$ & $0.8961 \pm 0.0153$ \\
        \texttt{Meta-FM} & $4.4171 \pm 0.2001$ & $0.5011 \pm 0.1171$ & $\underline{0.8975} \pm \underline{0.0086}$ \\
        \midrule
        \multicolumn{4}{l}{\textbf{\textit{Static (ours)}}} \\[1mm]
        \texttt{M2M-ED} & $4.5184 \pm 0.2323$ & $0.5689 \pm 0.1551$ & $0.8882 \pm 0.0101$ \\
        \texttt{M2M}-$\mathcal{W}_1$ & $\textbf{4.1005} \pm \textbf{0.2270}$ & $0.8270 \pm 0.1242$ & $0.7660 \pm 0.0228$ \\
        \texttt{M2M-OTMSE} & $\underline{4.2772} \pm \underline{0.1792}$ & $1.0085 \pm 0.0962$ & $0.7211 \pm 0.0210$ \\
        \midrule
        \multicolumn{4}{l}{\textbf{\textit{Dynamic (ours)}}} \\[1mm]
        \texttt{M2M-TFM} & $4.3526 \pm 0.2175$ & $\textbf{0.4715} \pm \textbf{0.1319}$ & $\textbf{0.9029} \pm \textbf{0.0050}$ \\
        \bottomrule
      \end{tabular}
    }
  \end{center}
  \vspace{-1.5em} %
\end{wraptable} 
We quantify prediction performance using distributional distances $\mathcal{W}_1, \text{ED}$, and $r^2$ (coefficient of determination) for each left-out test measure. 
Further experimental details are in \cref{ap:pdo_expt_details}. 

\textbf{Results.} \cref{tab:pdo_table} has results for the task of predicting patient-specific treatment response for unseen patients. 
Although each patient has numerous replicate experiments of pre- and post-treatment single-cell population (measure pairs), this task is especially difficult as measurements are only available for 10 patients in this dataset. 
As a result, generalizing to an unseen patient (model trained on data from 9 patients, and evaluated on the left-out patient) requires a model to learn the M2M operator from limited patient-level ``observations''.
We observe that generally our static approaches perform best on the metrics corresponding to their respective losses (i.e. $\text{ED}$-loss and $\mathcal{W}_1$-loss), while exhibiting worse performance on the counterpart metrics. 
In contrast, \tfm yields a balanced performance across all metrics. 
The strong $\mathcal{W}_1$ performance of \texttt{M2M}-$\mathcal{W}_1$ suggests that measure-dependence is beneficial here, 
while \tfm's balanced performance makes it the most robust choice in practice.

\vspace{-1.7em}
\section{Conclusion \& Discussion}\label{sec:conclusion}
\vspace{-1em}

In this work, we established transformers as a natural and powerful architecture for measure-to-measure (M2M) regression.  
By exploiting the inherent measure-dependence of transformers, we introduced both \textit{static} and \textit{dynamic} frameworks for learning nonlinear M2M regression operators. 
Notably, \tfm (our \textit{dynamic} approach) integrates flow matching with the measure-dependent nature of transformers, yielding competitive performance across a range of M2M regression tasks, including the challenging real-world task of predicting patient-specific treatment response. 
Together, our novel methods provide practical, scalable, and expressive solutions for M2M regression.

\paragraph{Limitations \& Future work} 
Our methods are currently limited to modeling simplified transport dynamics of physical processes. 
We only considered linear interpolation paths for our \textit{dynamic} M2M regression approach (i.e. \tfm), yet many physical processes, such as cell dynamics, inherently exhibit nonlinear, stochastic, and birth-death dynamics. 
Promising directions are to extend
\tfm to settings with non-gradient dynamics \citep{petrovic2026curly, rathod2026splineflowflowmatchingdynamical,guan2026call}, stochasticity \citep{tong2023simulation}, and the unbalanced setting \citep{peng2026wfr}.

Another avenue for future research is to use our static approaches as one-step generative models, in line with recent advances in this direction~\citep{deng2026drifting, zhu2026simple, geng2026mean, song2024improved, boffi2024flow, potaptchik2026meta}. Since our \textit{static} approaches already yield one-step transport maps from arbitrary source distributions, adapting our framework to generative modeling is straightforward.
Applying \tfm to higher-dimensional data, such as images, is also promising. \citet{luo2026flow} expand the source measure beyond a single Gaussian, a setting that our M2M framework naturally accommodates, while \citet{li2025back} show that transformers can be trained directly in pixel space.
Lastly, our framework could be extended to other large-scale perturbation screen datasets, such as \citet{zhang2025tahoe} and \citet{huang2025x}.

Beyond empirical extensions, addressing certain theoretical questions for \textit{dynamic} M2M regression with transformers remains an important open direction. 
For instance, future work could study the identifiability of nonlinear M2M operators learned via vector field models under our flow matching framework.
Ultimately, our work establishes two transformer-based approaches as a natural and scalable paradigm for M2M regression, providing a versatile foundation for these subsequent questions and future work.

\section*{Acknowledgments}

This research was funded by the Natural Sciences and Engineering Research Council of Canada (NSERC) and the Canada CIFAR AI Chairs program.
The research was also enabled by computational resources provided by the Digital Research Alliance of Canada (\url{https://alliancecan.ca}) and the Alberta Machine Intelligence Institute (\url{https://www.amii.ca/}). In addition, LA was partially supported by the Eric and Wendy Schmidt Center at the Broad Institute of MIT and Harvard, and by the NSERC Postdoctoral Fellowship.

\bibliographystyle{plainnat}
\bibliography{main}

\clearpage
\newpage

\appendix

\section{Additional Experimental Details}
\label{ap:expt_details}

\subsection{Multi-measure Objects} \label{ap:letters_expt_details}

In this section we outline additional experimental setup details for the Multi-measure Objects experiments. The kernel interaction process in this experiment is defined by the following dynamics equation, 
\[
x_{t+\Delta t} = x_t + \eta(x_t - \sum_j A_{ij} x_j)\Delta t + \sigma\sqrt{\Delta t}\epsilon
\]
where $A_{ij} = \exp(-\|x_i - x_j\|^2 / 2h^2) / \sum_k \exp(-\|x_i - x_k\|^2 / 2h^2)$ is the normalized Gaussian kernel with bandwidth $h$, $\eta$ is the repulsion strength, and $\sigma$ is the noise scale. For this experiment, we use $\eta=0.3$, $\sigma=0.001$, $h=0.75$, $\Delta t = 0.05$, and simulate the process for $50$ steps.

\subsection{McKean-Vlasov Systems}
\label{ap:mvsde_expt_details}

\subsubsection{Dynamics for McKean-Vlasov Systems}\label{app:dynamics_mkv}

We use \textit{modified} versions of the McKean-Vlasov Systems from \citet{pmlr_nmkv}, specifically the Kuramoto model, the Mean-Field Atlas system, and the FitzHugh-Nagumo model each of which we modify. Our goal in modifying these systems is to ensure that given different initial measures the resulting end states of the system are different. This was done because a naive extension of these systems to higher dimensions corresponded to lower diversity in the final states when changing the initial conditions. We enumerate the equations we used for governing a particular particle $i$ of the $N$ particles at time $t$, which we denote by $X_{i,t}$ as follows. $X_{i,t}$ is a $d$ dimensional vector representing one particle. Note that $X_{j,t}$ for a specific j denotes another particle in the system at time $t$, i.e $i \neq j$, and that $W_i$ is a Wiener random process, sampled individually for each $i$. Additionally we use $X_{i,t}^{{\text{dim }} d}$ to denote the $d$th dimension of a particle, i.e one of the entries in the $D$ dimensional vector.

We simulate the evolution of the Kuramoto system following equation \ref{eq:Kuramoto_System} with initial conditions consisting of $333$ particles drawn from $U_{[-1, 1]}$ and $167$ particles drawn from $\mathcal{N}(A, 0.1)$ with A sampled from $U_{[-1,1]}$. We use $\sigma = 0.2$ and we simulate $100$ time points from $t=0$ to $t=5.0$. 
\begin{align} \label{eq:Kuramoto_System}
    & dX_{i,t} = \left(\sin(X_{i,t}) + (2 / N) \sum_{j=1}^N \left(\sin(X_{j,t} - X_{i,t})\right) \right)dt + \sigma * dW_i
\end{align}  

For the modified FitzHugh-Nagumo system we use a different equation for the first dimension, $X_{i,t}^{1}$ and the other dimensions $X_{i,t}^2$. Dimension $1$ is simulated according to equation $\ref{eq:Fitz_Dim1}$, and all other dimensions are simulated using equation $\ref{eq:Fitz_Dim2}$. We simulate starting from initial conditions consisting of $333$ particles drawn from $U_{[-6,6]}$ and $167$ drawn from $\mathcal{N}(A, 0.1)$  with A drawn from $U_{[-6,6]}$. We use a different value of $b,c,d$ for each dimension. We set the values of each of these coefficients to be evenly spaced over the following ranges $b_i \in [0.5, 0.8], c_i \in [0.5, 1.0], d_i \in [0.3,0.7]$ and $\tau_i \in [1.0, 10.0]$. We simulate $100$ time-points evenly spaced between the initial time and final time. We note that for $D=2$ we simulate to time $4.0$ and for $D>2$ we simulate until time $10.0$. 

\begin{align}
     & \mathbf{b} = [b_1, \cdots, b_D], b_i = (0.5 + \frac{i-1}{D-1}0.3)) \hspace{4pt} \mathbf{c}  = [c_1, \cdots, c_D], c_i = (0.5 + \frac{i-1}{D-1}0.5)) \nonumber \\
     & \mathbf{d} = [d_1, \cdots, d_D], d_i = (0.3 + \frac{i-1}{D-1}0.4)) \hspace{4pt} \bm{\tau} = [
     \tau_1, \cdots, \tau_D], \tau_i = (1.0 + \frac{i-1}{D-1}9.0)) \nonumber \\
     & I(t) = 0.1 \times \sin(10 \times t) \nonumber \hspace{10pt} \lambda=0.4, a=0.2 \\
    & dX_{i,t}^{\text{dim} 1} = \left(a X^{\text{dim} 1}_{i,t}(X_{i,t}^{\text{dim} 1} - \lambda)(1-X^{\text{dim} 1}_{i,t}) - \frac{1}{N}\sum_{j=1}^n X^{\text{dim} d}_{j,t} + I(t) \right) dt \nonumber \\
    & \quad - \left( X_{i,t}^{\text{dim} 1} - \frac{1}{N}\sum_{j=1}^N X_{j,t}^{\text{dim} 1} \right) dt \label{eq:Fitz_Dim1}  \\
    & dX_{i,t}^{\text{dim} d} = \left(\frac{-\mathbf{b}^{
\text{dim} d} \times 
    X^{\text{dim} d}_{i,t} + \mathbf{c}^{
\text{dim} d} \times X^{\text{dim} 1}_{i,t} + \mathbf{d}^{
\text{dim} d}}{\bm{\tau}^{
\text{dim} d}} \right) dt \label{eq:Fitz_Dim2} 
\end{align}

For the MeanField Atlas we simulate using equation \ref{eq:Atlas}. With initial conditions consisting of $333$ particles drawn from $U_{[-2,2]}$ and $167$ drawn from $\mathcal{N}(A, 0.3)$ where $A$ is drawn from $U_{[-1,1]}$. We use $\sigma = 0.5$, and we simulate for $100$ time-points evenly spaced between 0 and $2.0$. 
\begin{align}
    & dX_{i,t}^{\text{dim } d} = \gamma\left(\frac{1}{N}\sum_{j=1}^N 1_{[x_{i,t} - X{j,t} \geq 0 ]}\right)*dt + \sigma * dW_{i,t} \label{eq:Atlas}\\ 
    & \quad \text{ with } \gamma(x) = 5 * (0.5 - x) + (X_{i,t} - 0.01 * X_{i,t}^3) + 1.5\sin(X_{i, t}^{\text{dim } d-1}) \nonumber
\end{align} 

Lastly we test on two additional and very simple systems which are easy to visualize and easy to infer the correct behaviour for. The first system we call GoToMean, and it attracts every particle to the mean of the system at each time point. We simulate using the dynamics in \cref{eq:GoToMean_System}. We simulate starting from initial conditions consisting of $334$ particles
drawn from $U_{[-10, 10]}$ and $166$ particles drawn from $\mathcal{N}(A, 0.1)$ with $A$ sampled from $U_{[-10,10]}$. This is a system without noise. We simulate $100$ time points from time $t=0$ to $t=10.0$. 
\begin{align} \label{eq:GoToMean_System} 
& dX_{i,t} = \left( (1 / N) \sum_{j=1}^N X_{j,t} - X_{i,t} \right) dt \end{align}
The second simple system we call GoToZero attracts every particle to origin at each time point. We simulate using the dynamics in \cref{eq:GoToZero_System}. We start initial conditions consisting of $334$ particles drawn from $U_{[-1, 1]}$ and $166$ particles drawn from
$\mathcal{N}(A, 0.01)$ with $A$ sampled from $U_{[-1,1]}$. We simulate $100$ time points from time $t=0$ to $t=10.0$. 
\begin{align} \label{eq:GoToZero_System}
    & dX_{i,t} = -X_{i,t} \, dt
\end{align} We provide results for these systems in \cref{app:additional_mkv}.

\subsubsection{Extra Training Details}

We shuffle data-points at each time point such that the datasets produced by simulating each of these systems do not have a one to one correspondence between particle i and index i at each time point. For training models on these systems with $100$ timepoints, we treat each pair of timepoints as if it were its own independent measure pair. To be specific, each measure trajectory is a 100 element tuple $(\mu_0, \mu_1, \cdots, \mu_{99})$, and we train our model by having it regress between $\mu_t, \mu_{t+1}$ independently. We generate these pairs to regress between by randomly sampling a batch of $16$ time-points from the tuple of length $100$. We update on each of these $16$ pairs in parallel. We choose to treat these pairs as independent rather than employ complex solutions used for multi-marginal flow matching \citep{Rohbeck2025ModelingCS}. This means that the models trained in this setup, sample two adjacent time-points for updating. Since the model is conditioned on time, it learns a single map from $\mu_0$ to $\mu_{99}$ formed by connecting all the learned maps between adjacent time points. Additionally since we have $500$ particles in each measure we sample a batch of $128$ particles at each training step. At inference time we only provide the models with the first time point, and simulate until the next time point, then we repeat this process using the output of the model as the next starting point. We do this until we reach the final time point to produce a predicted trajectory.  

\subsection{Patient-derived Organoid (PDO) Dataset}
\label{ap:pdo_expt_details}

We adopt the data processing framework established in \citet{atanackovic2024meta}. We refer the reader to \citet{atanackovic2024meta} (Appendix D.2) for a more detailed outline of the PDO dataset. 

\begin{figure*}[ht]
    \centering
    \begin{subfigure}[b]{0.32\textwidth}
        \centering
        \includegraphics[width=\textwidth]{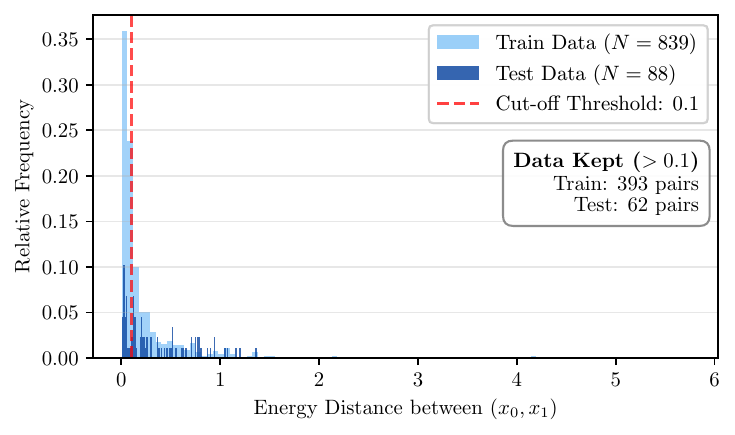}
        \caption{PDO-21}
        \label{fig:sub1}
    \end{subfigure}
    \hfill %
    \begin{subfigure}[b]{0.32\textwidth}
        \centering
        \includegraphics[width=\textwidth]{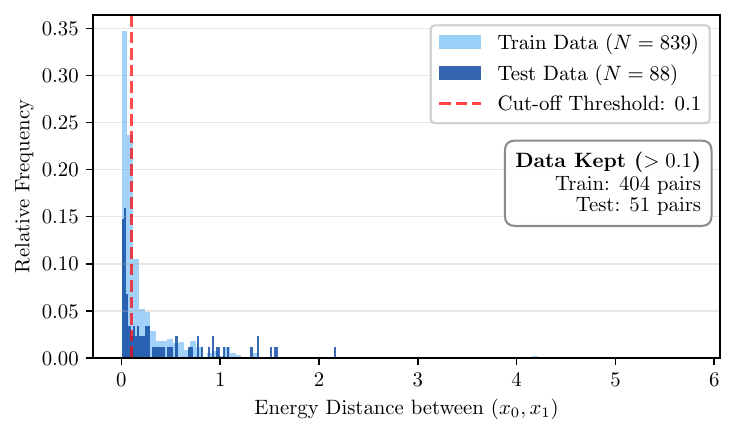}
        \caption{PDO-27}
        \label{fig:sub2}
    \end{subfigure}
    \hfill %
    \begin{subfigure}[b]{0.32\textwidth}
        \centering
        \includegraphics[width=\textwidth]{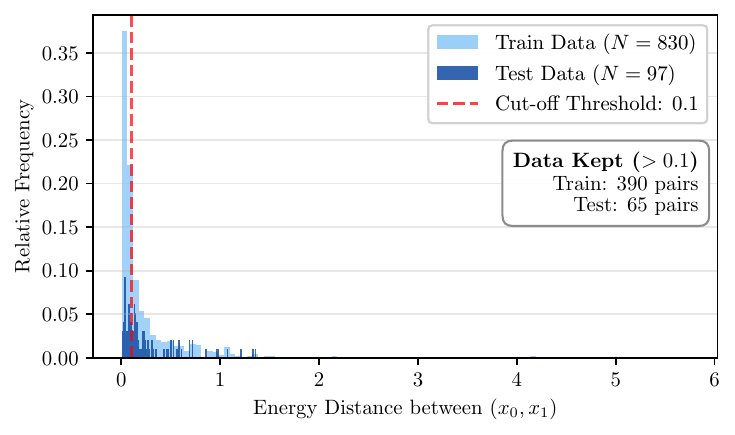}
        \caption{PDO-75}
        \label{fig:sub3}
    \end{subfigure}
    \caption{Empirical distribution of ED across all measures. Statistics reported for each split.}
    \label{fig:ed_pairs}
\end{figure*}

To ensure high-quality training data, we applied additional quality control (QC) filtering. 
Specifically, we select measure pairs with sufficiently large distributional distance (determined by some threshold). We achieve this by computing ED across all measure pairs in the dataset, and dropping all pairs that don't meet the threshold. We select a threshold of ED = 0.1. 
We show the empirical distribution of ED across all measures for the respective splits in \cref{fig:ed_pairs} and the corresponding split statistics.

Lastly, we pair all training cells in each training measure pair per-split pre-training. 
We do this by first projecting the data features (bio-markers) onto the top 5 principal components of the training data in each split. 
We then iterate through all measures and acquire the OT pairing between the respective measures.

Models are evaluated throughout training on $9$ randomly selected left-out measure pairs from the training set. The 1-Wasserstein metric is tracked every 50 epochs, and the best model checkpoint is selected based on this metric, and likewise used for evaluation on the test set (i.e. left out patient measure pairs).

\subsection{Additional Information on Baselines}\label{app:baselines}

In this section we offer information about our baseline methods. 
All models incorporate positional encoding for samples/particles. For time-dependent models, we use sinusoidal embeddings for time. 

\paragraph{Conditional Flow Matching (\texttt{CFM})} For our \texttt{CFM} baseline, we follow the setup from \cite{atanackovic2024meta}. For multi-measure objects experiments, we use a densely-connected MLP based architecture. 
For the McKean-Vlasov and biological experiments, we use a ResNet architecture.
For the biological experiments (\textbf{\textit{iii}}, \cref{sec:patients}), \citet{atanackovic2024meta} demonstrated that \texttt{OT-CFM} yields competitive performance relative to CellOT \citep{Bunne2023NeuralOT}, an input convex neural network (ICNN)-based approach, on the colorectal cancer treatment-response dataset, which we consider in \cref{sec:patients}. 
Similar to \citep{Bunne2023NeuralOT}, \texttt{OT-CFM} provides an optimal transport-based approximation for the M2M operator, and as a result, we use \texttt{OT-CFM} as a competitive baseline in the biological setting and do not compare directly to CellOT. 

\paragraph{Meta Flow Matching (\texttt{Meta-FM})} We follow the setup from \cite{atanackovic2024meta}, but forgo the graph convolutional network (GCN) distribution encoder and use a transformer as the encoding architecture instead. The transformer architecture similarly maps an empirical distribution, i.e a matrix input, into a single vector output. For the McKean-Vlasov experiments, we do follow the setup from \cite{atanackovic2024meta} and use the graph convolutional network (GCN).  
We use the same vector field model as in \texttt{CFM}), with the additional condition on the distribution embedding. We train \texttt{Meta-FM} for twice the gradient steps relative to \texttt{CFM} as it required alternating updates of two set of parameters. This provides a fair comparison with \texttt{CFM}. 

\paragraph{Wasserstein Flow Matching (\texttt{WFM})} We consider \texttt{WFM} as an additional baselines for the multi-measure objects experiments. 
We use the exact architecture made public in the code released with the original paper in \citep{haviv2024wasserstein}. 
\texttt{WFM} operates by first pairing measures (i.e. does not assume \textit{paired} measures) with a \textit{meta}-OT step (OT over measures).
We additionally added particle batching within each pair of measures to be consistent with our multi-measure objects experiment setup. 
We do this to ensure that \texttt{WFM} and the other methods train on roughly the same amount of data for a fair comparison. In our experiments we use $16$ measures as the batch of measures. 

\paragraph{Neural McKean Vlasov (\texttt{NMKV})} For the method from \citet{pmlr_nmkv}, \texttt{NMKV}, we use the empirical measure architecture with a modified training loop to introduce batching of particles (i.e. to match our setting). We set the gradient based learning of sigma to be false, since this is the default in \citet{pmlr_nmkv}. We do not enable Brownian bridge interpolation since our data does not contain missing/irregular time-points. We set the hidden dim for `MF model` to $0$ because we wish to use the empirical measure architecture which performed well in the original work. We use the implementation of their empirical measure and Mean Field MLP network architecture from the provided codebase \citep{pmlr_nmkv}. 

\paragraph{Mean Field Transformer (\texttt{MF-Transformer})} For the method from \citet{mean_field_biswal}, \texttt{MF-Transformer}, we use the published model, with one key modification. We add a sinusoidal timestep embedding, concatenation to the input, i.e appended to each particle, and passed into the model's embedding layer. The rest of the model is unchanged other than the embedding layer width increasing to handle the time concatenation. We do this because predicting the evolution of multi-time point McKean Vaslov systems requires knowing which of $99$ measure pairs the model is currently in. 
We train this model to regress towards their proposed finite-difference approximation target. The approximation target in the McKean Vlasov setting corresponds to the following, 
\[
\text{Target} := \frac{\left( \mu_{t+1} - \mu_{t-1}   \right)}{\left( 2 * \Delta t \right)}
\]
Notably the setup in this work assumes paired particles and we do not use paired particles, nor do we add OT coupling to this method as such it is expected that this method will struggle in this setting.

\subsection{Metrics and Distributional Losses}
\label{ap:metrics_and_losses}

We consider a variety of distributional distances to quantify model performance and to serve as loss functions for our static approaches. Specifically, we consider the 1-Wasserstein and 2-Wasserstein distances, the maximum mean discrepancy (MMD) distance, and energy distance (ED). We additionally consider mean squared error (MSE) as a loss function when used in conjunction with optimal transport couplings. We provide pseudo-code for training and inference of the \textit{static} method in \cref{alg:static_training} and \cref{alg:static_inference} respectively. 

\textbf{Maximum Mean Discrepancy (MMD) Distance and Energy Distance (ED).} Let $\hat{\mu} = \{x_1, \cdots, x_k\}$ and $\hat{\nu} = \{y_1, \cdots, y_m\}$ denote two empirical distributions with $x_i \in \mathbb{R}^d$ and $y_i \in \mathbb{R}^d$. For loss function, we always have $k = m$, while metrics do not require this constraint (predicted and target distributions may differ in the number of samples). We denote the maximum mean discrepancy distance as
\[ 
   \texttt{MMD}(\hat{\mu}, \hat{\nu}) := \frac{1}{k^2} \sum_{i=1}^k \sum_{i'=1}^k k(x_i, x_{i'}) + \frac{1}{m^2} \sum_{j=1}^m \sum_{j'=1}^m k(y_j, y_{j'}) - \frac{2}{km} \sum_{i=1}^k \sum_{j=1}^m k(x_i, y_j),
\]
where $k(\cdot, \cdot)$ is a kernel function. For what we label as ``maximum mean discrepancy (MMD)'', we use a Radial Basis Kernel for $k$ with parameter $\gamma$. When using MMD as a loss function (experiments in \cref{sec:objects}) we use $\gamma = 0.05$. When using MMD as a metric (in extended results in \cref{ap:letters_additional_expts}), we compute an average over $\gamma = [2, 1, 0.5, 0.1, 0.01, 0.005]$.
Energy distance (ED) then simply follows by consider the kernel $k(x, y) = - \Vert x - y \Vert$.

\textbf{Wasserstein Distance.} For $p \geq 1$, we can define the $p$-Wasserstein distance $(\mathcal{W}_p)$ as 
\[ %
    \mathcal{W}_p(\hat{\mu}, \hat{\nu}) := \left( \inf_{\gamma \in \Gamma(\mu, \nu)} \int \Vert (x - y) \Vert^p  d\gamma(x,y) \right)^{\frac{1}{p}},
\]
with $\Gamma$ denoting the set of couplings between $\mu$ and $\nu$, i.e the set of pairings between elements of $\mu$ to elements of $\nu$. For model performance comparisons, we evaluate the 1-Wasserstein distance ($\mathcal{W}_1$ for $p=1$) and the 2-Wasserstein distance ($\mathcal{W}_2$ for $p=2$).  

\textbf{Mean squared error (MSE).} We consider MSE, used in conjunction with mini-batch optimal transport, as an additional loss function. Given given coupled samples $(x, y) \sim \gamma(x,y)$ with $k = m = N$, we denote the \texttt{OTMSE} loss as
\begin{align*}
    & \texttt{MSE}_\gamma(\hat{\mu}, \hat{\nu}) := \frac{1}{N} \sum_{i=1}^N \Vert x_i - y_i \Vert ^2.
\end{align*}

\begin{figure} %
  \begin{minipage}{\linewidth} 
    \vspace{-2.2em} %
    
    \begin{algorithm}[H] 
      \caption{\textit{Static} M2M (\textit{Training step})}
      \label{alg:static_training}
      \begin{lstlisting}[belowskip=-0.2em, aboveskip=-0.05em, mathescape=true, basicstyle=\footnotesize\linespread{0.4}\selectfont\ttfamily]
# Inputs:
#  Dataset: $\color{pastel_green}\mathcal{M} = \{(\mu_i, \nu_i)\}_{i=1}^n$ 
#       with $\color{pastel_green}\mu_i \color{pastel_green}\in \mathbb{R}^{\color{pastel_green}N_{\mu_i} \times \color{pastel_green}d}$, $\color{pastel_green}\nu_i \color{pastel_green}\in \mathbb{R}^{\color{pastel_green}N_{\nu_i} \times \color{pastel_green}d}$
#  $\color{pastel_green}b$:(num measures) $\color{pastel_green}m$:(particles per measure) 
#  $\color{pastel_green}\theta$: Transformer parameters, $\color{pastel_green}F^\theta_\mu(\cdot)$

x, y = sample_measures($\mathcal{M}$, $b$, $m$) 
# tuple ([$\color{pastel_green}b, \color{pastel_green}m, \color{pastel_green}d$], [$\color{pastel_green}b, \color{pastel_green}m, \color{pastel_green}d$])  
 
# if using non-distributional loss add: 
# x, y = ot_couplings(x, y)  

y_pred = Transformer(x, $\theta$)

loss = D(y, y_pred)
$\theta$ = update(loss, $\theta$)
      \end{lstlisting}
    \end{algorithm}

    \vspace{-1.5em} %

    \begin{algorithm}[H]
      \caption{\textit{Static} M2M (\textit{Inference / sampling})}
      \label{alg:static_inference}
      \begin{lstlisting}[belowskip=-0.2em, aboveskip=-0.05em, mathescape=true,basicstyle=\footnotesize\linespread{0.4}\selectfont\ttfamily]
# Inputs:
#   Test Measure: $\color{pastel_green} \mu_j \color{pastel_green}\in \mathbb{R}^{\color{pastel_green}N_{\mu_j} \times \color{pastel_green}d}$ , 
#   $\color{pastel_green}\theta$: Transformer parameters, $\color{pastel_green}F^\theta_\mu(\cdot)$

x = $\mu_j$ # [$\color{pastel_green}1, \color{pastel_green}N_{\mu_j}, \color{pastel_green}d$]

y = Transformer(x, $\theta$)

return y # approximates true $\color{pastel_green}\nu_j$ 
      \end{lstlisting}
    \end{algorithm}
    
    \vspace{-3em} %
  \end{minipage}
\end{figure}
We do not compute MSE as a metric for model evaluation.

\textbf{Coefficient of determination ($r^2$).} This metric was considered by \citet{atanackovic2024meta} and \citet{Bunne2023NeuralOT} to evaluate method performance between model predicted treatment distributions and target treatment distributions. We use $r^2$ for our biological single-cell experiments in \cref{sec:patients}. In essence, this metric computes a correlation of correlations between predicted and target empirical distributions. We refer the reader to \citet{atanackovic2024meta} for a definition of this metric. 

\textbf{Cosine Distance. } Given two vectors $a,b$ we compute cosine distance as \[ d_{\cos}(a,b) \;=\; 1 - \frac{\langle a, b\rangle}{\|a\|_2\,\|b\|_2}.\] We use this in the McKean–Vlasov systems experiments in which we take the average cosine distance between two measures. Thus we have to apply it to two empirical measures, $A$ and $B$ each consisting of $n$ vectors $a_i$ and $b_i$. To do this we compute \[ d_{mean \cos}(a,b) \;=\; \frac{1}{n} \sum_{i=1}^n 1 - \frac{\langle a_i, b_i\rangle}{\|a_i\|_2\,\|b_i\|_2}.\]

\textbf{Computation of Losses.} To compute distributional losses efficiently, we employ the Geometric Loss Functions package \citep{feydy2019interpolating}. This package provides auto-differentiable versions of these distances enabling efficient and scalable training of our static approaches. Specifically for the $p$-Wasserstein loss, we approximate $\mathcal{W}_p$ (for $p=1$ and $p=2$) via the Sinkhorn algorithm \citep{cuturi2013sinkhorn}.

\textbf{Metrics.} We use these corresponding distributional distances as metrics to evaluate the performance of all methods. Since our objective of all methods is to recover the correct \textit{target} distribution, we do not consider MSE as a metric for evaluation.

\subsection{Architecture and Hyperparameters}
\label{app:arch_and_hparams}

\paragraph{Architecture}\label{app:arch}
Our architecture is a transformer model which accepts as input an empirical measure represented by a matrix of points. At a high level our transformer consists of an input projection, a sequence of residual blocks, and a final output projection. Our input projection begins by augmenting each point with sinusoidal spatial Fourier features (using a learned random matrix with 128 frequencies). This is then fed into an linear input projection layer to the hidden dimension size. Time is encoded and passed into the model with a sinusoidal timestep embedding. Together, this is then fed into a stack of AdaLNBlocks \citep{film_transformer} blocks, each block uses adaptive layer normalization modulated by time embedding (and any additional covariates/conditions that are used, e.g. treatments in the biological experiments), and provides gating. Between each block there is multi-head attention, which acts globally across all samples in the current measure. 
The final layer consists of a linear projection to the ambient dimension. 
Our architecture design is akin to transformer architectures used for image generation via diffusion and stochastic interpolants \citep{peebles2023scalable, ma2024sit}.

We use the same architecture for the static M2M methods but omit time conditioning, as it operates strictly as a one-step flow map. We optimize our model with Adam \citep{Kingma2014AdamAM} using its default hyperparameters. In \cref{app:ourmodel_hypers} and \cref{app:baseline_hypers} we list the hyperparameters used for our experiments. 

\paragraph{Hyperparameter Tuning for multi-measure objects} 
For the multi-measure objects experiment, we tuned hyperparameters of all methods with a basic grid search over small ranges. For our static and dynamic M2M approaches, we predominately tuned model width, depth, learning rate. 
We ablated width (number of hidden units = [64, 128, 256, 512]) and depth (number of transformer blocks = [1,2,3,4,5]), and learning rate = [5e-4, 1e-4, 5e-5].
We found performance of our static method predominantly depended on depth (i.e. number of transformer \textit{blocks}), while the dynamic approach was generally invariant. For \texttt{CFM}, we used the base hyperparamters from \cite{atanackovic2024meta}. We also used the base hyperparameters for the vector field in our \texttt{Meta-FM} implementation, but tuned our new self-attention-based distribution encoder using similar depth and width. 
We evaluated on the held-out ``X'' silhouettes to tune models, but remark, extensive tuning was not required to find a solid performing set of hyperparameters.

For \texttt{WFM} we initially used the default hyperparameters as given in the provided code from \citep{haviv2024wasserstein}, however we observed these hyperparameters led to over-fitting and so we lowered the number of training iterations, reduced the number of layers, and reduced the number of heads. We list these values in \cref{tab:wfm_unpaired_kernel_hparams}. We found that with a smaller network \texttt{WFM} performed better in this setting. 

\paragraph{Hyperparameter tuning for McKean-Vlasov experiments} We now discuss how we set the hyperparameters for our models, listed in, \cref{app:ourmodel_hypers} and for the baseline models, listed in \cref{app:baseline_hypers}. For the McKean-Vlasov experiments we performed a grid search learning rate sweep for all methods on a $100$ dimensional Kuramoto system using a held out seed, i.e a system not used for evaluating any of the models, and choose the best learning rate to apply to all other systems. We test the following learning rates $[0.001,0.0001,0.00001,0.000001]$, and for \texttt{Meta-FM} we test the cross product of this list for both the encoder and decoder learning rates. For \texttt{Meta-FM} we also tune the number of nearest neighbours $k \in [1, 10, 50]$. For learning rates which lead to divergence when applied to the other systems, we lowered the learning rates until no divergence occurred. For all other hyperparameters we attempt to ensure the models have similar number of parameters and hence we set the number of layers to $5$, and the hidden widths to $512$. All models which use dropout have it set to $0.1$.

\paragraph{Hyperparameter tuning for biological experiments}
For biological experiments, similar to letters, we conducted a minimal grid search over model depth, width, learning rate, and number of training epochs. Specifically, for our M2M methods, we ablated width (number of hidden units = [64, 128, 256, 512]) and depth (number of transformer blocks = [3, 5,  7, 9]), and learning rate = [5e-4, 1e-4, 5e-5].
We used a validation set of left out measure pairs within each of the 9 training patients for each respective split to evaluate performance throughout training, and evaluate on this validation set every 50 epochs. We used the best performing checkpoint found throughout training on this validation set as the final model for inference on the test set (completely unseen measures from the left-out patient). 

\subsection{Hyperparameters used by our approaches}\label{app:ourmodel_hypers}
 For the multi-measure object experiments in \cref{sec:objects} \tfm uses the hyperparameters listed in \cref{tab:letters_simple_attention_flow_hparams}. For the multi-time-point McKean Vlasov experiments in \cref{sec:exp_mkvs} \tfm uses the hyperparameters listed in \cref{tab:m2m_v2_hparams}. For the patient-derived organoid dataset experiments in \cref{sec:patients} \tfm uses the hyperparameters listed in \cref{tab:trellis_simple_attention_flow_hparams}. The hyperparameters used by the static approach for the multi-measure object experiments in \cref{sec:objects} are listed in table \cref{tab:static_multi_objects}. For the McKean Vlasov experiments they are listed in \cref{tab:sweep_map_hparams}, and for derived organoid dataset experiments  they are listed in \cref{tab:static_prediction}.  For the McKean Vaslov system experiment, unless otherwise written all methods use a linear learning rate schedule with a start factor $1.0$ and an end factor of $0.01$.
 
\begin{table}[ht]
\centering
\begin{minipage}[t]{0.45\textwidth}
   \caption{Hyperparameters for the \tfm model on multi-measure Object experiments.}
    \centering
    \begin{tabular}[t]{l c}
    \hline
    \textbf{Hyperparameter}                 & \textbf{Value} \\
    \hline
    Input dimension                         & 2 \\
    Hidden dimension                        & 64 \\
    Number of layers                        & 5 \\
    Number of attention heads               & 8 \\
    Learning rate                           & $5 \times 10^{-5}$ \\
    Dropout                                 & 0.05 \\
    Time embedding dimension                & 128 \\
    Time-varying dynamics                   & True \\
    Time steps at inference                 & 100 \\
    Measure Batch Size                      & 16 \\
    Particle Batch Size                     & 512 \\
    OT during Training                      & True \\
    Epochs (1 training pass over dataset)   & 1000 \\
    \hline
    \end{tabular}
 
    \label{tab:letters_simple_attention_flow_hparams}
\end{minipage}
\hfill %
\begin{minipage}[t]{0.45\textwidth}
    \caption{Hyperparameters for the \tfm model for the Patients experiments.}
    \centering
    \begin{tabular}[t]{l c}
    \hline
    \textbf{Hyperparameter}                      & \textbf{Value} \\
    \hline
    Input dimension                              & 44 \\
    Hidden dimension                             & 512 \\
    Number of layers                         & 5 \\
    Number of attention heads                    & 8 \\
    Learning rate                                & $1 \times 10^{-4}$ \\
    Dropout                                      & 0.05 \\
    Time steps at inference                       & 100 \\
    Measure Batch Size                              & 1 \\
    Particle Batch Size                              & 2048 \\
    OT during Training                           & False \\
    Epochs       & 750 \\
    \hline
    \end{tabular}
    \label{tab:trellis_simple_attention_flow_hparams}
\end{minipage}
\end{table}

\begin{table}[H]
\centering
\caption{Hyperparameters for the \tfm transformer model in McKean Vlasov Experiments}
\begin{tabular}{l c}
\hline
\textbf{Hyperparameter}              & \textbf{Value} \\
\hline
Hidden dimension                     & 512 \\
Number of layers                     & 5 \\
Number of attention heads            & 4 \\
Learning Rate                        & $1 \times 10^{-4}$ \\
Dropout                              & 0.1 \\
Time embedding dimension             & 128 \\
Steps between marginals at inference & 100 \\
Measure Batch Size                 & 16 \\
Particle Batch Size                 & 128 \\
OT during Training                   & True  \\
Iterations                          & 100,000 \\
\hline
\end{tabular}
\label{tab:m2m_v2_hparams}
\end{table}

\begin{table}[H]
\centering
\begin{minipage}[t]{0.45\textwidth}
    \centering
    \caption{Hyperparameters for \texttt{M2M-MMD}, \texttt{M2M-ED}, \texttt{M2M-}$\mathcal{W}_2$, \texttt{M2M-}$\mathcal{W}_1$ and \texttt{M2M-OTMSE} models on multi-measure object experiments.}
    \begin{tabular}[t]{l c}
    \hline
    \textbf{Hyperparameter}                 & \textbf{Value} \\
    \hline
    Input dimension                         & 2 \\
    Number of layers                        & 5 \\
    Number of attention heads               & 8 \\
    Hidden dimension                        & 64 \\
    Learning rate                           & $1 \times 10^{-4}$ \\
    Dropout                                 & 0.05 \\
    Time steps at inference                 & 1 \\    
    Measure Batch Size                      & 16 \\
    Particle Batch Size                      & 512 \\
     OT during Training                      & True \\
    Epochs (1 training pass over dataset)   & 2000 \\
    \hline
    \end{tabular}
    \label{tab:static_multi_objects}
\end{minipage}
\hfill
\begin{minipage}[t]{0.45\textwidth}
    \centering
    \caption{Hyperparameters for \texttt{M2M-ED}, \texttt{M2M-OTMSE} and \texttt{M2M-}$\mathcal{W}_1$ in McKean-Vlasov Experiments.}
    \begin{tabular}[t]{l c}
    \hline
    \textbf{Hyperparameter}                  & \textbf{Value} \\
    \hline
    Learning rate                            & $1 \times 10^{-4}$ \\
    Number of layers                         & 5 \\
    Number of attention heads                & 4 \\
    Hidden dimension                         & 512 \\
    Dropout                                  & 0.1 \\
    Time embedding dimension                 & 128 \\
    Training iterations                      & 100,000 \\
    Steps per marginal at inference          & 1 \\
    Measure Batch Size                       & 16 \\
    Particle Batch Size                      & 128 \\
    OT during Training                       & True \\
    \hline
    \end{tabular}
    \label{tab:sweep_map_hparams}
\end{minipage}
\end{table}

\begin{table}[H]
\vspace{-1em}
\centering
    \caption{Hyperparameters for  \texttt{M2M-MMD}, \texttt{M2M-ED},  \texttt{M2M-}$\mathcal{W}_2$,  \texttt{M2M-}$\mathcal{W}_1$ and \texttt{M2M-OTMSE} in the biological experiments.}
    \centering
    \begin{tabular}[t]{l c}
    \hline
    \textbf{Hyperparameter}                      & \textbf{Value} \\
    \hline
    Input dimension                              & 44 \\
    Hidden dimension                             & 512 \\
    Number of layers                             & 7 \\
    Number of attention heads                    & 8 \\
    Learning rate                                & $1 \times 10^{-4}$ \\
    Dropout                                      & 0.05 \\
    Time steps at inference                      & 1 \\
    Measure Batch size                           & 1 \\
    Particle Batch Size                          & 2048 \\
    OT during Training                           & False \\
    Epochs                                       & 1000 \\
    \hline
    \end{tabular}
    \label{tab:static_prediction}
    \vspace{-1.5em}
\end{table}

\subsection{Baseline Models Hyperparameters}\label{app:baseline_hypers}
For the multi-measure objects experiments we list the \texttt{CFM} model hyperparameters in \cref{tab:fm_letters_hparams}. We use the same  \texttt{Meta-FM} hyperparameters as \citep{atanackovic2024meta}, but instead of a graph convolution network we use a transformer population embedding which uses 3 layers, with a width of 64, and $8$ heads. The hyperparameters for  \texttt{Meta-FM} are shown in table \cref{tab:mfm_letters_hparams}. For \texttt{WFM} the hyperparameters used are in \cref{tab:wfm_unpaired_kernel_hparams}. Note that we used the default hyperparameters plus our batching and we found that the model overfit, as a result we used less gradient steps $(20,000)$ and with less layers $(4)$ which led to improved performance. For the McKean-Vlasov experiments in \cref{sec:exp_mkvs} we list hyperparameters in \cref{tab:fm_hparams} for \texttt{CFM}, \cref{tab:neuralmkv_hparams} for \texttt{NMKV}, \cref{tab:meantrans_hparams} for \texttt{MF-Transformer}, and \cref{tab:mfm_hparams} for \texttt{Meta-FM}. We note that for \texttt{CFM}, \tfm, and all static transformer methods a linear learning rate schedule was used with a start factor $1.0$ and an end factor of $0.01$ where each training step the learning rate was decayed until reaching the final factor at the end of training. For the patients experiments in \cref{sec:patients}, we list the \texttt{CFM} hyperparameters in \cref{tab:fm_trellis_hparams}, and  the hyperparameters for \texttt{Meta-FM} in \cref{tab:mfm_trellis_hparams}. For the McKean Vaslov system experiment, unless otherwise written all methods use a linear learning rate schedule with a start factor $1.0$ and an end factor of $0.01$.

\begin{table}[H]
\centering
\begin{minipage}[t]{0.48\textwidth}
    \centering
    \caption{Hyperparameters for the \texttt{CFM} model in the multi-measure object experiments.}
    \begin{tabular}[t]{l c}
        \hline
        \textbf{Hyperparameter} & \textbf{Value} \\
        \hline
        Input dimension                         & 2 \\
        Hidden dimension                        & 512 \\
        Number of layers                        & 4 \\
        Skip connections                        & False \\
        OT during Training                      & True \\
        Time steps at inference                 & 100 \\
        Learning rate                           & $1 \times 10^{-4}$ \\
        Measure Batch Size                      & 16 \\
        Particle Batch Size                     & 512 \\
        Epochs (passes over training data)      & 1000 \\
        \hline
    \end{tabular}
    \label{tab:fm_letters_hparams}
\end{minipage}
\hfill %
\begin{minipage}[t]{0.48\textwidth}
    \centering
    \caption{Hyperparameters for \texttt{Meta-FM} model in the multi-measure object experiments.}
    \begin{tabular}[t]{l c}
        \hline
        \textbf{Hyperparameter} & \textbf{Value} \\
        \hline
        Decoder Width                           & 512 \\
        Number of decoder layers                & 4 \\
        Transformer hidden dimension            & 64 \\
        Number of transformer layers            & 3 \\
        Number of attention heads               & 8 \\
        Flow learning rate                      & $1 \times 10^{-4}$ \\
        Encoder learning rate                   & $1 \times 10^{-4}$ \\
        OT during Training                      & True \\
        Steps at Inference                      & 100 \\
        Measure Batch Size                      & 1 \\
        Particle Batch Size                     & 2048 \\
        Epochs (passes over dataset)            &  2000 \\
        \hline
    \end{tabular}
    \label{tab:mfm_letters_hparams}
\end{minipage}
\end{table}

\begin{table}[H]
\centering
\begin{minipage}[t]{0.48\textwidth}
\caption{Hyperparameters for \texttt{CFM} model in McKean-Vlasov experiments.}
    \centering
    \resizebox{\textwidth}{!}{%
    \begin{tabular}{l c}
    \hline
\textbf{Hyperparameter}              & \textbf{Value} \\
\hline
Hidden dimension                     & 512 \\
Number of layers                     & 5 \\
Learning Rate                        & $1 \times 10^{-6}$ \\
Dropout                              & No \\
Time embedding dimension             & 128 \\
Steps per marginal at inference & 100 \\
Measure Batch Size                   & 16 \\
Particle Batch Size                 & 128 \\
OT during Training                   & True  \\
Iterations  & 100,000 \\
    \hline
    \end{tabular}
    }
    \label{tab:fm_hparams}
\end{minipage}
\hfill %
\begin{minipage}[t]{0.48\textwidth}
    \centering
    \caption{Hyperparameters for \texttt{Meta-FM} \citep{atanackovic2024meta} model in McKean-Vlasov experiments.}
    \resizebox{\textwidth}{!}{%
    \begin{tabular}{l c}
        \hline
        \textbf{Hyperparameter} & \textbf{Value} \\
        \hline
        Hidden dimension  Decoder                   & 512 \\
Hidden dimension  Encoder                   & 512 \\
Number of layers  & 5 \\
Number of nearest neighbours & 50 \\
Learning Rate Decoder                       & $1 \times 10^{-4}$ \\
Learning Rate Encoder                       & $1 \times 10^{-6}$ \\
Dropout                              & No \\
Time embedding dimension             & 128 \\
Steps per marginal at inference & 100 \\
Measure Batch Size     & 16 \\
Particle Batch Size                & 128 \\
OT during Training                   & True  \\
Iterations  & 200,000 \\
        \hline
    \end{tabular}
    }
    \label{tab:mfm_hparams}
\end{minipage}
\end{table}

\begin{table}[H]
\centering
\begin{minipage}[t]{0.45\textwidth}
\caption{Hyperparameters for \texttt{MF-Transformer} \citep{mean_field_biswal} model in McKean-Vlasov experiment.}
    \centering
     \resizebox{\textwidth}{!}{%
    \begin{tabular}[t]{l c}
        \hline
        \textbf{Hyperparameter} & \textbf{Value} \\
        \hline
        Hidden dimension                     & 512 \\
        Number of layers                     & 5 \\
        Number of attention heads            & 4 \\
        Learning Rate                        & $1 \times 10^{-6}$ \\
        Cosine annealing LR schedule         &$\eta_{\min} = 10^{-7}$ \\
        Dropout                              & No \\
        Time embedding dimension             & 128 \\
        Steps per marginal at inference      & 100 \\
        Measure Batch Size                   & 16 \\
        Particle Batch Size                  & 128 \\
        OT during Training                   & False  \\
        Iterations (1 gradient step)         & 100,000 \\
        \hline
            \end{tabular}}    
    \label{tab:meantrans_hparams}
\end{minipage}
\hfill %
\begin{minipage}[t]{0.45\textwidth}
    \centering
    \caption{Hyperparameters for \texttt{NMKV} \citep{pmlr_nmkv} model in McKean-Vlasov experiments.}
    \begin{tabular}[t]{l c}
    \hline
    \textbf{Hyperparameter} & \textbf{Value} \\
    \hline
    Hidden dimension                   & 512 \\
Number of layers                     & 5 \\
Learning Rate                        & $1 \times 10^{-5}$ \\
Exponential LR schedule & $\gamma = 0.9998$ per step \\
Particle Batch Size                  & 128 \\
Measure Batch Size                   & 16 \\
OT during Training                   & False  \\
Model Sigma                          & 1.0 \\
Number of Epochs                     & 100,000 \\
    \hline
    \end{tabular} %
    \label{tab:neuralmkv_hparams}
\end{minipage}
\end{table}

\begin{table}[H]
\centering
\begin{minipage}[t]{0.48\textwidth}
\caption{Hyperparameters for \texttt{CFM} in the biological experiments}
    \centering
    \begin{tabular}[t]{l c}
    \hline
    \textbf{Hyperparameter}                      & \textbf{Value} \\
    \hline
    Input dimension                              & 44 \\
    Width                                       & 512 \\
    Number of flow layers                       & 7 \\
    Learning rate                               & $1 \times 10^{-4}$ \\
    OT during Training                          & False \\
    Steps at Inference                           & 100 \\
    Measure Batch Size                           & 1 \\
    Particle Batch Size                          & 2048 \\
    Epochs (passes over the data)                & 750 \\
    \hline
    \end{tabular}
    \label{tab:fm_trellis_hparams}
\end{minipage}
\hfill %
\begin{minipage}[t]{0.48\textwidth}
    \centering
     \caption{Hyperparameters for \texttt{Meta-FM} model in the biological experiment.}
    \begin{tabular}[t]{l c}
    \hline
    \textbf{Hyperparameter}                 & \textbf{Value} \\
    \hline
    Input dimension                         & 44 \\
    Decoder hidden dimension                & 512 \\
    Number of decoder layers                & 7 \\
    Transformer hidden dimension            & 64 \\
    Number of transformer layers            & 3 \\
    Number of attention heads               & 8 \\
    Flow learning rate                      & $1 \times 10^{-4}$ \\
    Encoder learning rate                   & $1 \times 10^{-4}$ \\
    OT during Training                      & False \\
    Steps at Inference                      & 100 \\
    Batch size                              & 1 \\
    Particle Batch Size                     & 2048 \\
    Epochs (passes over data)               & 1500 \\
    Validation frequency (epochs)           & 250 \\
    \hline
    \end{tabular}
    \label{tab:mfm_trellis_hparams}
\end{minipage}
\end{table}

\begin{table}[H]
\centering
\caption{Hyperparameters for \texttt{WFM} \citep{haviv2024wasserstein} in the multi-measure object experiments.}
\begin{tabular}{l c}
\hline
\textbf{Hyperparameter} & \textbf{Value} \\
\hline
Input dimension & 2 \\
Number of transformer layers & 4 \\
Number of attention heads & 4 \\
Point embedding dimension & 256 \\
MLP hidden dimension & 256 \\
Dropout rate & 0.1 \\
Monge map / OT coupling & Rounded matching \\
Sinkhorn iterations & 200 \\
Entropic OT regularization $\varepsilon$ & $2 \times 10^{-3}$ \\
Data scaling & None \\
Learning rate (initial) & $2 \times 10^{-4}$ \\
LR schedule & Exponential decay (every $1000$ steps, factor $0.998$) \\
Training iterations (gradient steps) & 20,000 (100, 000 found to do worse) \\
Measure Batch Size                          & 16 \\
Particle Batch Size                          & 512 \\
Steps at inference  & 100 \\
\hline
\end{tabular}
\label{tab:wfm_unpaired_kernel_hparams}
\end{table}

\section{Computational Requirements}\label{app:comp_used}
We run our experiments using a research compute cluster with L$40$ Nvidia GPUs (48GB) and H$100$ GPUs (80 GB). All jobs utilized GPUs, with CPU demand being negligible.  All jobs took between $2-24$ hours to complete. The multi-measure objects experiment took $2$ to $4$ hours on a single L$40$, using the entire GPU memory. The McKean-Vlasov experiments were run on $1/4$ of an L$40$ using NVIDIA Multi-Instance GPU, this means they had access to $1/4$ the compute and $1/4$ the memory of an L$40$. A single seed used approximately $6$ hours of time to complete, with negligible variance between different models. Lastly the patients experiments used a single $L40$ and took approximately $8$ hours again with small variance between models.

\section{Additional Experiments and Results}
\label{ap:additional_expts}

\subsection{Multi-measure Objects} \label{ap:letters_additional_expts}

\begin{figure*}[t]
    \vspace{-2em}
    \centering
    \begin{subfigure}[b]{0.46\textwidth}
        \centering
        \includegraphics[width=\linewidth]{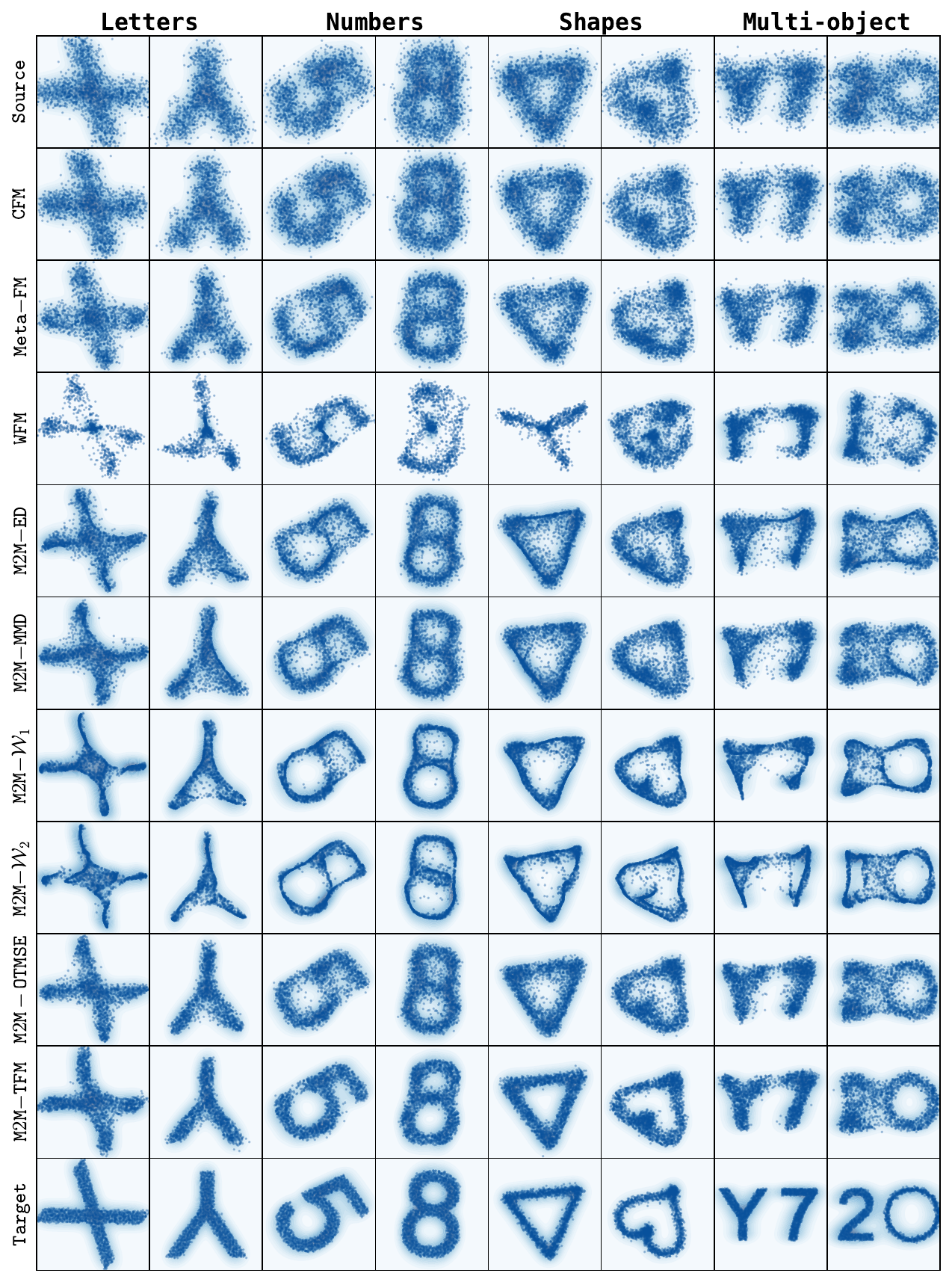}
        \caption{\textbf{Diffusion process}}
        \label{fig:diffusion_plot_app}
    \end{subfigure}
    \hspace{0.05\textwidth}
    \begin{subfigure}[b]{0.46\textwidth}
        \centering
        \includegraphics[width=\linewidth]{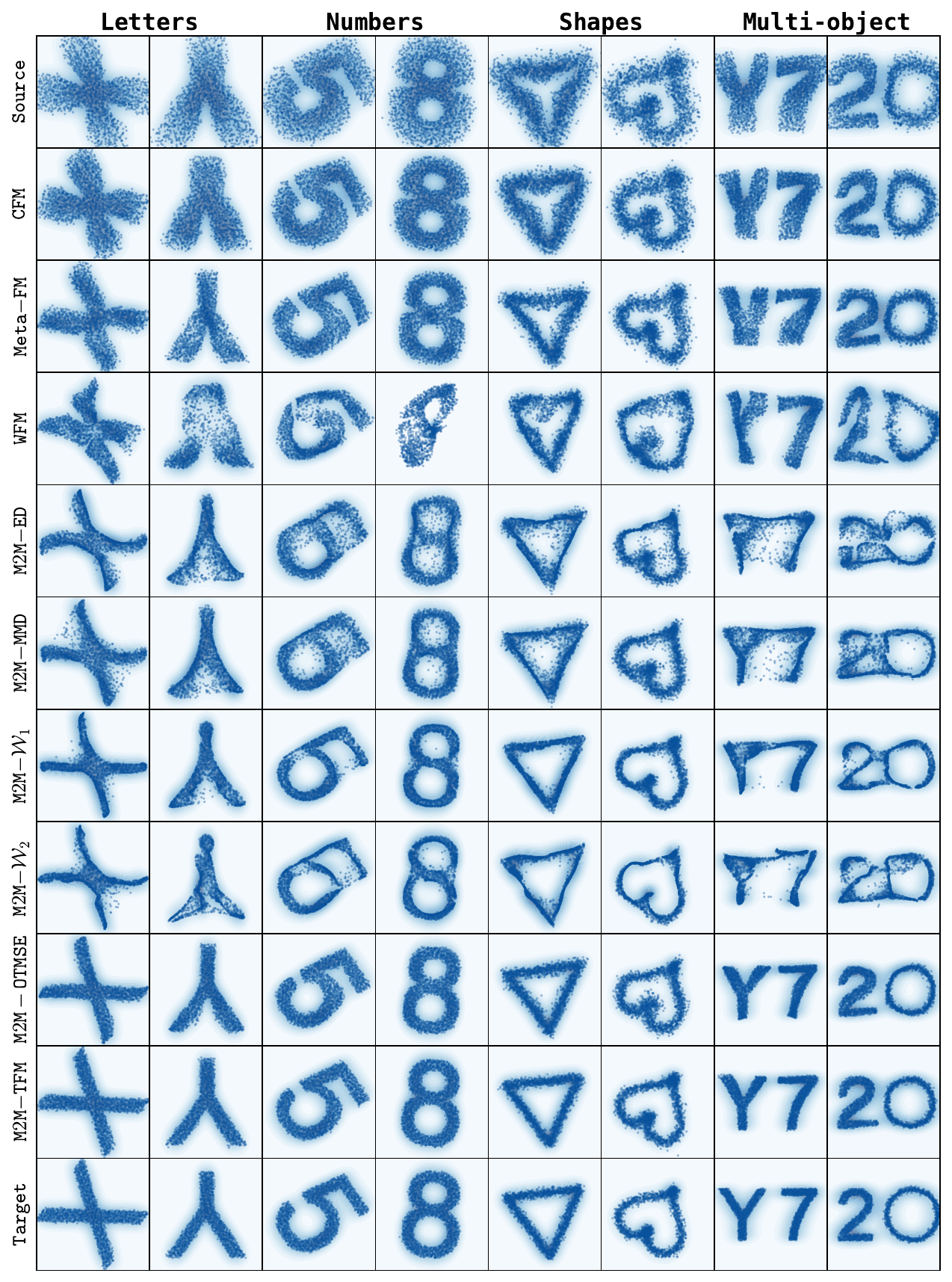}
        \caption{\textbf{Kernel interaction process}}
        \label{fig:kernel_plot_app}
    \end{subfigure}
    
    \caption{Extended visualization of model predictions on multi-measure objects for unseen measures.}
    \label{fig:letters_viz_extended}
\end{figure*}

We present extended results for the multi-measure objects experiments in \cref{tab:letters_extended_diffusion_comparison} and \cref{tab:letters_extended_kernel_interactions_comparison}, as well as extended visuals in \cref{fig:letters_viz_extended}.

\begin{table}[t]
\caption{Prediction performance for \textbf{Diffusion} system. Mean $\pm$ Std computed over 5 random seeds. \textbf{Bold} indicates the best performing method.}
\label{tab:letters_extended_diffusion_comparison}
\begin{center}
\begin{small}
\begin{sc}
\resizebox{\columnwidth}{!}{%
\begin{tabular}{llcccc}
\toprule[1.2pt]
 & \texttt{\textbf{Model}} & $\mathcal{W}_1$ ($\downarrow$) & $\mathcal{W}_2$ ($\downarrow$) & MMD ($\downarrow$) & ED ($\downarrow$) \\
\midrule[1pt]
\multirow{9}{*}{\textbf{\texttt{Letters}}} & $\mathtt{CFM}$ & $0.2407 \pm 0.0022$ & $0.3137 \pm 0.0028$ & $0.0032 \pm 0.0001$ & $0.0066 \pm 0.0003$ \\
 & $\mathtt{Meta}{-}\mathtt{FM}$ & $0.2146 \pm 0.0170$ & $0.2715 \pm 0.0193$ & $0.0027 \pm 0.0005$ & $0.0070 \pm 0.0019$ \\
 & $\mathtt{WFM}$ & $0.6299 \pm 0.1538$ & $0.7228 \pm 0.1615$ & $0.0424 \pm 0.0207$ & $0.1200 \pm 0.0563$ \\
 & $\mathtt{M2M}{-}\mathtt{ED}$ & $0.1819 \pm 0.0116$ & $0.2385 \pm 0.0147$ & $0.0020 \pm 0.0003$ & $0.0048 \pm 0.0006$ \\
 & $\mathtt{M2M}{-}\mathtt{MMD}$ & $0.1876 \pm 0.0246$ & $0.2386 \pm 0.0284$ & $0.0021 \pm 0.0005$ & $0.0073 \pm 0.0031$ \\
 & $\mathtt{M2M}{-}\mathcal{W}_1$ & $0.2027 \pm 0.0068$ & $0.2418 \pm 0.0123$ & $0.0026 \pm 0.0006$ & $0.0081 \pm 0.0010$ \\
 & $\mathtt{M2M}{-}\mathcal{W}_2$ & $0.2243 \pm 0.0201$ & $0.2684 \pm 0.0229$ & $0.0033 \pm 0.0008$ & $0.0089 \pm 0.0013$ \\
 & $\mathtt{M2M}{-}\mathtt{OTMSE}$ & $0.1339 \pm 0.0054$ & $0.1676 \pm 0.0054$ & $0.0010 \pm 0.0002$ & $0.0027 \pm 0.0005$ \\
 & $\mathtt{M2M}{-}\mathtt{TFM}$ & $\mathbf{0.1118 \pm 0.0035}$ & $\mathbf{0.1361 \pm 0.0042}$ & $\mathbf{0.0006 \pm 0.0001}$ & $\mathbf{0.0017 \pm 0.0001}$ \\
\midrule
\multirow{9}{*}{\textbf{\texttt{Numbers}}} & $\mathtt{CFM}$ & $0.2060 \pm 0.0047$ & $0.2710 \pm 0.0056$ & $0.0021 \pm 0.0001$ & $0.0045 \pm 0.0004$ \\
 & $\mathtt{Meta}{-}\mathtt{FM}$ & $0.1895 \pm 0.0085$ & $0.2403 \pm 0.0106$ & $0.0019 \pm 0.0002$ & $0.0051 \pm 0.0012$ \\
 & $\mathtt{WFM}$ & $0.6478 \pm 0.3541$ & $0.7665 \pm 0.3824$ & $0.0557 \pm 0.0666$ & $0.1561 \pm 0.1847$ \\
 & $\mathtt{M2M}{-}\mathtt{ED}$ & $0.1714 \pm 0.0124$ & $0.2230 \pm 0.0153$ & $0.0014 \pm 0.0003$ & $0.0038 \pm 0.0008$ \\
 & $\mathtt{M2M}{-}\mathtt{MMD}$ & $0.1527 \pm 0.0084$ & $0.2047 \pm 0.0090$ & $0.0010 \pm 0.0001$ & $0.0037 \pm 0.0012$ \\
 & $\mathtt{M2M}{-}\mathcal{W}_1$ & $0.1721 \pm 0.0054$ & $0.2152 \pm 0.0091$ & $0.0017 \pm 0.0003$ & $0.0050 \pm 0.0007$ \\
 & $\mathtt{M2M}{-}\mathcal{W}_2$ & $0.1971 \pm 0.0194$ & $0.2408 \pm 0.0225$ & $0.0022 \pm 0.0005$ & $0.0067 \pm 0.0018$ \\
 & $\mathtt{M2M}{-}\mathtt{OTMSE}$ & $0.1376 \pm 0.0068$ & $0.1837 \pm 0.0071$ & $0.0008 \pm 0.0001$ & $0.0022 \pm 0.0005$ \\
 & $\mathtt{M2M}{-}\mathtt{TFM}$ & $\mathbf{0.1072 \pm 0.0037}$ & $\mathbf{0.1306 \pm 0.0045}$ & $\mathbf{0.0004 \pm 0.0000}$ & $\mathbf{0.0012 \pm 0.0003}$ \\
\midrule
\multirow{9}{*}{\textbf{\texttt{Shapes}}} & $\mathtt{CFM}$ & $0.2689 \pm 0.0061$ & $0.3324 \pm 0.0076$ & $0.0041 \pm 0.0003$ & $0.0090 \pm 0.0009$ \\
 & $\mathtt{Meta}{-}\mathtt{FM}$ & $0.2516 \pm 0.0286$ & $0.3083 \pm 0.0289$ & $0.0041 \pm 0.0010$ & $0.0109 \pm 0.0039$ \\
 & $\mathtt{WFM}$ & $0.5900 \pm 0.2258$ & $0.6744 \pm 0.2383$ & $0.0570 \pm 0.0533$ & $0.1541 \pm 0.1361$ \\
 & $\mathtt{M2M}{-}\mathtt{ED}$ & $0.2026 \pm 0.0072$ & $0.2507 \pm 0.0067$ & $0.0026 \pm 0.0002$ & $0.0069 \pm 0.0007$ \\
 & $\mathtt{M2M}{-}\mathtt{MMD}$ & $0.2082 \pm 0.0145$ & $0.2578 \pm 0.0165$ & $0.0028 \pm 0.0004$ & $0.0075 \pm 0.0015$ \\
 & $\mathtt{M2M}{-}\mathcal{W}_1$ & $0.1951 \pm 0.0122$ & $0.2377 \pm 0.0126$ & $0.0028 \pm 0.0004$ & $0.0082 \pm 0.0014$ \\
 & $\mathtt{M2M}{-}\mathcal{W}_2$ & $0.2138 \pm 0.0261$ & $0.2505 \pm 0.0309$ & $0.0035 \pm 0.0013$ & $0.0107 \pm 0.0046$ \\
 & $\mathtt{M2M}{-}\mathtt{OTMSE}$ & $0.1780 \pm 0.0020$ & $0.2231 \pm 0.0029$ & $0.0019 \pm 0.0000$ & $0.0047 \pm 0.0002$ \\
 & $\mathtt{M2M}{-}\mathtt{TFM}$ & $\mathbf{0.1286 \pm 0.0041}$ & $\mathbf{0.1558 \pm 0.0057}$ & $\mathbf{0.0009 \pm 0.0001}$ & $\mathbf{0.0024 \pm 0.0001}$ \\
\midrule
\multirow{9}{*}{\textbf{\texttt{Multi-object}}} & $\mathtt{CFM}$ & $0.2806 \pm 0.0018$ & $0.3519 \pm 0.0025$ & $0.0035 \pm 0.0001$ & $0.0080 \pm 0.0002$ \\
 & $\mathtt{Meta}{-}\mathtt{FM}$ & $0.2666 \pm 0.0046$ & $0.3299 \pm 0.0043$ & $0.0040 \pm 0.0002$ & $0.0098 \pm 0.0005$ \\
 & $\mathtt{WFM}$ & $0.4660 \pm 0.0571$ & $0.5478 \pm 0.0654$ & $0.0157 \pm 0.0057$ & $0.0505 \pm 0.0199$ \\
 & $\mathtt{M2M}{-}\mathtt{ED}$ & $0.2577 \pm 0.0078$ & $0.3334 \pm 0.0088$ & $0.0036 \pm 0.0004$ & $0.0094 \pm 0.0010$ \\
 & $\mathtt{M2M}{-}\mathtt{MMD}$ & $0.2659 \pm 0.0067$ & $0.3489 \pm 0.0211$ & $0.0038 \pm 0.0001$ & $0.0108 \pm 0.0006$ \\
 & $\mathtt{M2M}{-}\mathcal{W}_1$ & $0.2670 \pm 0.0123$ & $0.3361 \pm 0.0148$ & $0.0047 \pm 0.0006$ & $0.0127 \pm 0.0013$ \\
 & $\mathtt{M2M}{-}\mathcal{W}_2$ & $0.2800 \pm 0.0199$ & $0.3510 \pm 0.0311$ & $0.0049 \pm 0.0007$ & $0.0143 \pm 0.0027$ \\
 & $\mathtt{M2M}{-}\mathtt{OTMSE}$ & $0.2226 \pm 0.0062$ & $0.2755 \pm 0.0069$ & $0.0026 \pm 0.0002$ & $0.0079 \pm 0.0009$ \\
 & $\mathtt{M2M}{-}\mathtt{TFM}$ & $\mathbf{0.1724 \pm 0.0022}$ & $\mathbf{0.2172 \pm 0.0021}$ & $\mathbf{0.0015 \pm 0.0001}$ & $\mathbf{0.0038 \pm 0.0003}$ \\
\midrule
\multirow{9}{*}{\textbf{\texttt{AVERAGE}}} & $\mathtt{CFM}$ & $0.2491 \pm 0.0032$ & $0.3173 \pm 0.0044$ & $0.0032 \pm 0.0001$ & $0.0071 \pm 0.0004$ \\
 & $\mathtt{Meta}{-}\mathtt{FM}$ & $0.2306 \pm 0.0114$ & $0.2875 \pm 0.0131$ & $0.0032 \pm 0.0003$ & $0.0082 \pm 0.0017$ \\
 & $\mathtt{WFM}$ & $0.5834 \pm 0.1399$ & $0.6779 \pm 0.1493$ & $0.0427 \pm 0.0305$ & $0.1201 \pm 0.0813$ \\
 & $\mathtt{M2M}{-}\mathtt{ED}$ & $0.2034 \pm 0.0073$ & $0.2614 \pm 0.0081$ & $0.0024 \pm 0.0002$ & $0.0062 \pm 0.0005$ \\
 & $\mathtt{M2M}{-}\mathtt{MMD}$ & $0.2036 \pm 0.0094$ & $0.2625 \pm 0.0088$ & $0.0024 \pm 0.0002$ & $0.0073 \pm 0.0012$ \\
 & $\mathtt{M2M}{-}\mathcal{W}_1$ & $0.2092 \pm 0.0055$ & $0.2577 \pm 0.0037$ & $0.0029 \pm 0.0001$ & $0.0085 \pm 0.0005$ \\
 & $\mathtt{M2M}{-}\mathcal{W}_2$ & $0.2288 \pm 0.0151$ & $0.2777 \pm 0.0189$ & $0.0035 \pm 0.0006$ & $0.0102 \pm 0.0023$ \\
 & $\mathtt{M2M}{-}\mathtt{OTMSE}$ & $0.1680 \pm 0.0032$ & $0.2125 \pm 0.0034$ & $0.0016 \pm 0.0001$ & $0.0044 \pm 0.0003$ \\
 & $\mathtt{M2M}{-}\mathtt{TFM}$ & $\mathbf{0.1300 \pm 0.0009}$ & $\mathbf{0.1599 \pm 0.0012}$ & $\mathbf{0.0009 \pm 0.0000}$ & $\mathbf{0.0023 \pm 0.0001}$ \\
\bottomrule[1.2pt]
\end{tabular}%
}
\end{sc}
\end{small}
\end{center}
\end{table}

\begin{table}[t]
\caption{Prediction performance for \textbf{Kernel interactions} system. Mean $\pm$ Std computed over 5 random seeds. \textbf{Bold} indicates the best performing method.}
\label{tab:letters_extended_kernel_interactions_comparison}
\begin{center}
\begin{small}
\begin{sc}
\resizebox{\columnwidth}{!}{%
\begin{tabular}{llcccc}
\toprule[1.2pt]
 & \texttt{\textbf{Model}} & $\mathcal{W}_1$ ($\downarrow$) & $\mathcal{W}_2$ ($\downarrow$) & MMD ($\downarrow$) & ED ($\downarrow$) \\
\midrule[1pt]
\multirow{9}{*}{\textbf{\texttt{Letters}}} & $\mathtt{CFM}$ & $0.3145 \pm 0.0048$ & $0.3630 \pm 0.0054$ & $0.0067 \pm 0.0002$ & $0.0146 \pm 0.0007$ \\
 & $\mathtt{Meta}{-}\mathtt{FM}$ & $0.2276 \pm 0.0209$ & $0.2716 \pm 0.0236$ & $0.0046 \pm 0.0009$ & $0.0128 \pm 0.0026$ \\
 & $\mathtt{WFM}$ & $0.5802 \pm 0.1671$ & $0.6877 \pm 0.1763$ & $0.0373 \pm 0.0242$ & $0.1078 \pm 0.0699$ \\
 & $\mathtt{M2M}{-}\mathtt{ED}$ & $0.1708 \pm 0.0413$ & $0.2237 \pm 0.0571$ & $0.0019 \pm 0.0008$ & $0.0054 \pm 0.0021$ \\
 & $\mathtt{M2M}{-}\mathtt{MMD}$ & $0.1525 \pm 0.0418$ & $0.2005 \pm 0.0610$ & $0.0015 \pm 0.0006$ & $0.0051 \pm 0.0023$ \\
 & $\mathtt{M2M}{-}\mathcal{W}_1$ & $0.1374 \pm 0.0306$ & $0.1679 \pm 0.0440$ & $0.0011 \pm 0.0003$ & $0.0039 \pm 0.0012$ \\
 & $\mathtt{M2M}{-}\mathcal{W}_2$ & $0.1662 \pm 0.0578$ & $0.2049 \pm 0.0761$ & $0.0023 \pm 0.0023$ & $0.0061 \pm 0.0052$ \\
 & $\mathtt{M2M}{-}\mathtt{OTMSE}$ & $0.0943 \pm 0.0060$ & $0.1096 \pm 0.0082$ & $0.0008 \pm 0.0002$ & $0.0025 \pm 0.0006$ \\
 & $\mathtt{M2M}{-}\mathtt{TFM}$ & $\mathbf{0.0569 \pm 0.0082}$ & $\mathbf{0.0652 \pm 0.0089}$ & $\mathbf{0.0003 \pm 0.0001}$ & $\mathbf{0.0008 \pm 0.0002}$ \\
\midrule
\multirow{9}{*}{\textbf{\texttt{Numbers}}} & $\mathtt{CFM}$ & $0.2336 \pm 0.0025$ & $0.2847 \pm 0.0035$ & $0.0036 \pm 0.0001$ & $0.0095 \pm 0.0004$ \\
 & $\mathtt{Meta}{-}\mathtt{FM}$ & $0.1889 \pm 0.0079$ & $0.2206 \pm 0.0122$ & $0.0025 \pm 0.0003$ & $0.0076 \pm 0.0014$ \\
 & $\mathtt{WFM}$ & $0.5811 \pm 0.1631$ & $0.6940 \pm 0.1718$ & $0.0443 \pm 0.0315$ & $0.1266 \pm 0.0875$ \\
 & $\mathtt{M2M}{-}\mathtt{ED}$ & $0.1460 \pm 0.0158$ & $0.1870 \pm 0.0301$ & $0.0012 \pm 0.0002$ & $0.0035 \pm 0.0004$ \\
 & $\mathtt{M2M}{-}\mathtt{MMD}$ & $0.1284 \pm 0.0240$ & $0.1666 \pm 0.0365$ & $0.0010 \pm 0.0003$ & $0.0032 \pm 0.0012$ \\
 & $\mathtt{M2M}{-}\mathcal{W}_1$ & $0.1163 \pm 0.0217$ & $0.1436 \pm 0.0292$ & $0.0008 \pm 0.0002$ & $0.0026 \pm 0.0004$ \\
 & $\mathtt{M2M}{-}\mathcal{W}_2$ & $0.1432 \pm 0.0427$ & $0.1796 \pm 0.0566$ & $0.0013 \pm 0.0006$ & $0.0039 \pm 0.0017$ \\
 & $\mathtt{M2M}{-}\mathtt{OTMSE}$ & $0.0960 \pm 0.0152$ & $0.1145 \pm 0.0140$ & $0.0007 \pm 0.0001$ & $0.0025 \pm 0.0007$ \\
 & $\mathtt{M2M}{-}\mathtt{TFM}$ & $\mathbf{0.0509 \pm 0.0052}$ & $\mathbf{0.0604 \pm 0.0072}$ & $\mathbf{0.0002 \pm 0.0001}$ & $\mathbf{0.0006 \pm 0.0001}$ \\
\midrule
\multirow{9}{*}{\textbf{\texttt{Shapes}}} & $\mathtt{CFM}$ & $0.2510 \pm 0.0036$ & $0.2944 \pm 0.0045$ & $0.0050 \pm 0.0002$ & $0.0111 \pm 0.0007$ \\
 & $\mathtt{Meta}{-}\mathtt{FM}$ & $0.2201 \pm 0.0312$ & $0.2556 \pm 0.0333$ & $0.0048 \pm 0.0011$ & $0.0131 \pm 0.0046$ \\
 & $\mathtt{WFM}$ & $0.5058 \pm 0.1400$ & $0.5663 \pm 0.1490$ & $0.0366 \pm 0.0339$ & $0.0973 \pm 0.0803$ \\
 & $\mathtt{M2M}{-}\mathtt{ED}$ & $0.1584 \pm 0.0161$ & $0.2004 \pm 0.0246$ & $0.0019 \pm 0.0002$ & $0.0050 \pm 0.0008$ \\
 & $\mathtt{M2M}{-}\mathtt{MMD}$ & $0.1555 \pm 0.0145$ & $0.1946 \pm 0.0197$ & $0.0019 \pm 0.0002$ & $0.0052 \pm 0.0010$ \\
 & $\mathtt{M2M}{-}\mathcal{W}_1$ & $0.1561 \pm 0.0173$ & $0.1932 \pm 0.0315$ & $0.0019 \pm 0.0006$ & $0.0062 \pm 0.0015$ \\
 & $\mathtt{M2M}{-}\mathcal{W}_2$ & $0.1695 \pm 0.0176$ & $0.2077 \pm 0.0281$ & $0.0023 \pm 0.0004$ & $0.0064 \pm 0.0008$ \\
 & $\mathtt{M2M}{-}\mathtt{OTMSE}$ & $0.1410 \pm 0.0094$ & $0.1702 \pm 0.0124$ & $0.0022 \pm 0.0004$ & $0.0059 \pm 0.0015$ \\
 & $\mathtt{M2M}{-}\mathtt{TFM}$ & $\mathbf{0.0774 \pm 0.0064}$ & $\mathbf{0.0907 \pm 0.0082}$ & $\mathbf{0.0007 \pm 0.0001}$ & $\mathbf{0.0018 \pm 0.0003}$ \\
\midrule
\multirow{9}{*}{\textbf{\texttt{Multi-object}}} & $\mathtt{CFM}$ & $0.2983 \pm 0.0051$ & $0.3436 \pm 0.0062$ & $0.0056 \pm 0.0002$ & $0.0160 \pm 0.0006$ \\
 & $\mathtt{Meta}{-}\mathtt{FM}$ & $0.2801 \pm 0.0212$ & $0.3280 \pm 0.0257$ & $0.0065 \pm 0.0009$ & $0.0186 \pm 0.0032$ \\
 & $\mathtt{WFM}$ & $0.4232 \pm 0.0768$ & $0.4864 \pm 0.0899$ & $0.0128 \pm 0.0045$ & $0.0407 \pm 0.0177$ \\
 & $\mathtt{M2M}{-}\mathtt{ED}$ & $0.2634 \pm 0.0411$ & $0.3299 \pm 0.0628$ & $0.0045 \pm 0.0011$ & $0.0133 \pm 0.0033$ \\
 & $\mathtt{M2M}{-}\mathtt{MMD}$ & $0.2535 \pm 0.0477$ & $0.3150 \pm 0.0602$ & $0.0043 \pm 0.0010$ & $0.0136 \pm 0.0034$ \\
 & $\mathtt{M2M}{-}\mathcal{W}_1$ & $0.2553 \pm 0.0366$ & $0.3150 \pm 0.0551$ & $0.0047 \pm 0.0010$ & $0.0150 \pm 0.0023$ \\
 & $\mathtt{M2M}{-}\mathcal{W}_2$ & $0.2763 \pm 0.0406$ & $0.3430 \pm 0.0678$ & $0.0051 \pm 0.0017$ & $0.0155 \pm 0.0033$ \\
 & $\mathtt{M2M}{-}\mathtt{OTMSE}$ & $0.2022 \pm 0.0103$ & $0.2253 \pm 0.0102$ & $0.0030 \pm 0.0001$ & $0.0103 \pm 0.0010$ \\
 & $\mathtt{M2M}{-}\mathtt{TFM}$ & $\mathbf{0.1268 \pm 0.0099}$ & $\mathbf{0.1455 \pm 0.0128}$ & $\mathbf{0.0014 \pm 0.0003}$ & $\mathbf{0.0046 \pm 0.0009}$ \\
\midrule
\multirow{9}{*}{\textbf{\texttt{AVERAGE}}} & $\mathtt{CFM}$ & $0.2744 \pm 0.0030$ & $0.3214 \pm 0.0041$ & $0.0052 \pm 0.0002$ & $0.0128 \pm 0.0005$ \\
 & $\mathtt{Meta}{-}\mathtt{FM}$ & $0.2292 \pm 0.0150$ & $0.2690 \pm 0.0162$ & $0.0046 \pm 0.0005$ & $0.0130 \pm 0.0022$ \\
 & $\mathtt{WFM}$ & $0.5226 \pm 0.0545$ & $0.6086 \pm 0.0494$ & $0.0328 \pm 0.0127$ & $0.0931 \pm 0.0303$ \\
 & $\mathtt{M2M}{-}\mathtt{ED}$ & $0.1847 \pm 0.0283$ & $0.2352 \pm 0.0431$ & $0.0024 \pm 0.0005$ & $0.0068 \pm 0.0014$ \\
 & $\mathtt{M2M}{-}\mathtt{MMD}$ & $0.1725 \pm 0.0312$ & $0.2192 \pm 0.0437$ & $0.0022 \pm 0.0005$ & $0.0068 \pm 0.0018$ \\
 & $\mathtt{M2M}{-}\mathcal{W}_1$ & $0.1663 \pm 0.0235$ & $0.2049 \pm 0.0351$ & $0.0021 \pm 0.0004$ & $0.0069 \pm 0.0009$ \\
 & $\mathtt{M2M}{-}\mathcal{W}_2$ & $0.1888 \pm 0.0380$ & $0.2338 \pm 0.0539$ & $0.0027 \pm 0.0011$ & $0.0080 \pm 0.0025$ \\
 & $\mathtt{M2M}{-}\mathtt{OTMSE}$ & $0.1334 \pm 0.0088$ & $0.1549 \pm 0.0095$ & $0.0017 \pm 0.0002$ & $0.0053 \pm 0.0009$ \\
 & $\mathtt{M2M}{-}\mathtt{TFM}$ & $\mathbf{0.0780 \pm 0.0020}$ & $\mathbf{0.0904 \pm 0.0029}$ & $\mathbf{0.0007 \pm 0.0001}$ & $\mathbf{0.0020 \pm 0.0002}$ \\
\bottomrule[1.2pt]
\end{tabular}%
}
\end{sc}
\end{small}
\end{center}
\end{table}

\begin{table}[t]
\caption{Ablation over the number of samples in each measure pair on the multi-measure objects experiment. Mean $\pm$ Std computed over 3 random model seeds and 8 ``\textit{object}'' silhouettes. \textbf{Bold} indicates the best performing method for each number of samples.}
\label{tab:letters_num_samples}
\begin{center}
\begin{small}
\begin{tabular}{lcccc}
\toprule
 & \multicolumn{2}{c}{\textbf{\texttt{Diffusion}}} & \multicolumn{2}{c}{\textbf{\texttt{Kernel interactions}}} \\
\cmidrule(lr){2-3} \cmidrule(lr){4-5}
\texttt{\textbf{Method}} & $\mathcal{W}_1$ ($\downarrow$) & ED ($\downarrow$) & $\mathcal{W}_1$ ($\downarrow$) & ED ($\downarrow$) \\
\midrule
\multicolumn{5}{l}{\textbf{\textit{512 samples}}} \\[1mm]
\texttt{CFM} & $0.2445 \pm 0.0011$ & $0.0066 \pm 0.0002$ & $0.2774 \pm 0.0010$ & $0.0133 \pm 0.0005$ \\
\texttt{Meta-FM} & $0.2449 \pm 0.0048$ & $0.0104 \pm 0.0003$ & $0.2074 \pm 0.0079$ & $0.0094 \pm 0.0011$ \\
\texttt{M2M-}$\mathcal{W}_1$ & $0.2015 \pm 0.0180$ & $0.0067 \pm 0.0014$ & $0.1033 \pm 0.0321$ & $0.0025 \pm 0.0009$ \\
\tfm & $\mathbf{0.1567 \pm 0.0024}$ & $\mathbf{0.0038 \pm 0.0000}$ & $\mathbf{0.0617 \pm 0.0023}$ & $\mathbf{0.0013 \pm 0.0001}$ \\
\midrule
\multicolumn{5}{l}{\textbf{\textit{1024 samples}}} \\[1mm]
\texttt{CFM} & $0.2470 \pm 0.0020$ & $0.0070 \pm 0.0003$ & $0.2771 \pm 0.0020$ & $0.0131 \pm 0.0004$ \\
\texttt{Meta-FM} & $0.2246 \pm 0.0123$ & $0.0079 \pm 0.0014$ & $0.2199 \pm 0.0076$ & $0.0119 \pm 0.0015$ \\
\texttt{M2M-}$\mathcal{W}_1$ & $0.1836 \pm 0.0193$ & $0.0059 \pm 0.0016$ & $0.1351 \pm 0.0405$ & $0.0050 \pm 0.0021$ \\
\tfm & $\mathbf{0.1310 \pm 0.0015}$ & $\mathbf{0.0023 \pm 0.0001}$ & $\mathbf{0.0748 \pm 0.0024}$ & $\mathbf{0.0018 \pm 0.0001}$ \\
\midrule
\multicolumn{5}{l}{\textbf{\textit{2048 samples}}} \\[1mm]
\texttt{CFM} & $0.2456 \pm 0.0009$ & $0.0068 \pm 0.0001$ & $0.2746 \pm 0.0027$ & $0.0127 \pm 0.0005$ \\
\texttt{Meta-FM} & $0.2254 \pm 0.0010$ & $0.0074 \pm 0.0004$ & $0.2325 \pm 0.0177$ & $0.0132 \pm 0.0020$ \\
\texttt{M2M-}$\mathcal{W}_1$ & $0.2030 \pm 0.0150$ & $0.0080 \pm 0.0016$ & $0.1803 \pm 0.0104$ & $0.0078 \pm 0.0017$ \\
\tfm & $\mathbf{0.1324 \pm 0.0022}$ & $\mathbf{0.0022 \pm 0.0001}$ & $\mathbf{0.0788 \pm 0.0076}$ & $\mathbf{0.0018 \pm 0.0003}$ \\
\midrule
\multicolumn{5}{l}{\textbf{\textit{4096 samples}}} \\[1mm]
\texttt{CFM} & $0.2481 \pm 0.0015$ & $0.0072 \pm 0.0000$ & $0.2755 \pm 0.0003$ & $0.0131 \pm 0.0002$ \\
\texttt{Meta-FM} & $0.2218 \pm 0.0068$ & $0.0073 \pm 0.0010$ & $0.2231 \pm 0.0086$ & $0.0114 \pm 0.0008$ \\
\texttt{M2M-}$\mathcal{W}_1$ & $0.1917 \pm 0.0250$ & $0.0074 \pm 0.0022$ & $0.1604 \pm 0.0555$ & $0.0068 \pm 0.0034$ \\
\tfm & $\mathbf{0.1282 \pm 0.0002}$ & $\mathbf{0.0020 \pm 0.0000}$ & $\mathbf{0.0835 \pm 0.0023}$ & $\mathbf{0.0021 \pm 0.0002}$ \\
\bottomrule
\end{tabular}
\end{small}
\end{center}
\end{table}

\begin{table}[t]
\caption{Unbalanced numbers of source and target samples. $\mathcal{W}_1$ ($\downarrow$) between predicted and target measures for models trained with different numbers of samples in the source and target measures. \textbf{Bold} indicates the best performing method for each setting.}
\label{tab:letters_unbalanced_w1}
\begin{center}
\begin{footnotesize}
\setlength{\tabcolsep}{4pt}
\begin{tabular}{cccccccccc}
\toprule
\multicolumn{2}{c}{\textbf{\texttt{\# samples}}} & \multicolumn{4}{c}{\textbf{\texttt{Diffusion}}} & \multicolumn{4}{c}{\textbf{\texttt{Kernel interactions}}} \\
\cmidrule(lr){1-2} \cmidrule(lr){3-6} \cmidrule(lr){7-10}
\texttt{\textbf{Source}} & \texttt{\textbf{Target}} & \texttt{CFM} & \texttt{Meta-FM} & \texttt{M2M-}$\mathcal{W}_1$ & \tfm & \texttt{CFM} & \texttt{Meta-FM} & \texttt{M2M-}$\mathcal{W}_1$ & \tfm \\
\midrule
1000 & 2000 & $0.2443$ & $0.2295$ & $0.2402$ & $\mathbf{0.1458}$ & $0.2760$ & $0.2337$ & $0.1939$ & $\mathbf{0.1068}$ \\
 & 4000 & $0.2459$ & $0.2693$ & $0.2062$ & $\mathbf{0.1675}$ & $0.2757$ & $0.2616$ & $0.1763$ & $\mathbf{0.1267}$ \\
 & 6000 & $0.2459$ & $0.2366$ & $0.1890$ & $\mathbf{0.1618}$ & $0.2735$ & $0.2293$ & $0.2180$ & $\mathbf{0.1375}$ \\
\midrule
3000 & 4000 & $0.2454$ & $0.2192$ & $0.1693$ & $\mathbf{0.1271}$ & $0.2784$ & $0.2383$ & $0.1075$ & $\mathbf{0.0850}$ \\
 & 6000 & $0.2453$ & $0.2235$ & $0.2148$ & $\mathbf{0.1244}$ & $0.2773$ & $0.2311$ & $0.2055$ & $\mathbf{0.0813}$ \\
\midrule
5000 & 6000 & $0.2443$ & $0.2369$ & $0.1718$ & $\mathbf{0.1247}$ & $0.2723$ & $0.2503$ & $0.1725$ & $\mathbf{0.0834}$ \\
\bottomrule
\end{tabular}
\end{footnotesize}
\end{center}
\end{table}

\subsection{McKean-Vlasov Systems} \label{app:additional_mkv}

We include tables with the metrics for all 9 systems, this includes the $2, 50,$ and $100$ dimension versions of our modified Kuramoto, Atlas, and FitzHugh-Nagumo systems. The results for the $2$ dimensional systems are listed in \cref{tab:2_mkv_results}, for the $50$ dimensional systems in \cref{tab:50_mkv_results}, and for $100$ dimensional systems in \cref{tab:100_mkv_results}. In \cref{tab:steps_mkv} we provide results for our ablation over the number of inference steps, i.e number of function evaluations used in inference for \tfm on the $100$ dimensional Kuramoto system. Lastly we provide a cosine distance metric between the true vector field and the learned vector field by the model. We provide these metrics for a subset of models on the $2, 50$ and $100$ dimensional systems in table \ref{tab:cosine_dim2}
, table \ref{tab:cosine_dim50}, and table \ref{tab:cosine_dim100}. As defined in \cref{ap:metrics_and_losses} the cosine reported here is the average cosine distance between all particles in two measures. This is further averaged over all $100$ time-points in the system. To compute this we query the model with the true marginal and compare the returned vector field to the true vector field that we use for simulation. This produces $100$ mean cosine distances and we take the average over all of them for our final value. Lastly we provide performance metrics for the two simple systems described in \cref{app:dynamics_mkv}, we provide these metrics in tables \cref{tab:extra_dim2}, \cref{tab:extra_dim50} and \cref{tab:extra_dim100}. We observe that for the GoToZero system, which does not require measure-dependence, that the non-measure dependent baseline often outperforms or performs on par with \tfm, this is in contrast to the GoToMean system where \tfm and the static M2M methods do generally outperform.

\begin{table}[ht]
\centering
{\caption{\textbf{Prediction performance comparison on $\mathbf{2}$-dimensional McKean-Vlasov systems.} Mean$\pm$Std computed over $6$ random seeds. \textbf{Bold} indicates the best performing method.}}
\resizebox{\textwidth}{!}{%
\begin{tabular}{lcccccc}
\toprule
 & \multicolumn{2}{c}{Kuramoto} & \multicolumn{2}{c}{FitzHugh-Nagumo} & \multicolumn{2}{c}{Atlas} \\
Method & $\mathcal{W}_1$ & $ED$ & $\mathcal{W}_1$ & $ED$ & $\mathcal{W}_1$ & $ED$ \\
\midrule
\texttt{CFM} & $1.454 \pm 0.021$ & $0.944 \pm 0.024$ & $0.381 \pm 0.001$ & $0.122 \pm 0.002$ & $1.400 \pm 0.012$ & $0.488 \pm 0.009$ \\
\texttt{MF-Transformer} & $1.450 \pm 0.251$ & $2.399 \pm 0.486$ & $1.291 \pm 0.073$ & $0.960 \pm 0.117$ & $1.230 \pm 0.012$ & $0.612 \pm 0.026$ \\
\texttt{NMKV} & $0.714 \pm 0.027$ & $0.952 \pm 0.051$ & $1.116 \pm 0.002$ & $0.762 \pm 0.003$ & $3.067 \pm 0.056$ & $3.805 \pm 0.112$ \\
\texttt{Meta-FM} & $0.285 \pm 0.016$ & $\mathbf{0.211 \pm 0.025}$ & $0.312 \pm 0.006$ & $0.080 \pm 0.003$ & $0.808 \pm 0.054$ & $0.285 \pm 0.044$ \\
\texttt{M2M-ED} & $0.502 \pm 0.025$ & $0.716 \pm 0.049$ & $0.386 \pm 0.033$ & $0.172 \pm 0.036$ & $0.892 \pm 0.107$ & $0.539 \pm 0.133$ \\
\texttt{M2M-OTMSE} & $0.319 \pm 0.029$ & $0.391 \pm 0.055$ & $0.344 \pm 0.019$ & $0.145 \pm 0.019$ & $0.977 \pm 0.059$ & $0.685 \pm 0.087$ \\
\texttt{M2M}-$\mathcal{W}_1$ & $0.373 \pm 0.026$ & $0.482 \pm 0.043$ & $0.355 \pm 0.030$ & $0.164 \pm 0.034$ & $0.900 \pm 0.068$ & $0.549 \pm 0.083$ \\
\tfm & $\mathbf{0.252 \pm 0.022}$ & $0.269 \pm 0.040$ & $\mathbf{0.288 \pm 0.007}$ & $\mathbf{0.070 \pm 0.006}$ & $\mathbf{0.678 \pm 0.035}$ & $\mathbf{0.225 \pm 0.026}$ \\
\bottomrule
\end{tabular}}
\label{tab:2_mkv_results}
\end{table}

\begin{table}[ht]
\centering
{\caption{\textbf{Prediction performance comparison on $\mathbf{50}$-dimensional McKean-Vlasov systems.} Mean$\pm$Std computed over $6$ random seeds. \textbf{Bold} indicates the best performing method.}}
\resizebox{\textwidth}{!}{%
\begin{tabular}{lcccccc}
\toprule
 & \multicolumn{2}{c}{Kuramoto} & \multicolumn{2}{c}{FitzHugh-Nagumo} & \multicolumn{2}{c}{Atlas} \\
Method & $\mathcal{W}_1$ & $ED$ & $\mathcal{W}_1$ & $ED$ & $\mathcal{W}_1$ & $ED$ \\
\midrule
\texttt{CFM} & $8.292 \pm 0.056$ & $5.200 \pm 0.070$ & $8.885 \pm 0.027$ & $0.341 \pm 0.008$ & $14.165 \pm 0.047$ & $4.514 \pm 0.013$ \\
\texttt{MF-Transformer} & $3.744 \pm 0.096$ & $2.262 \pm 0.193$ & $14.939 \pm 0.145$ & $1.855 \pm 0.041$ & $11.502 \pm 0.023$ & $5.517 \pm 0.032$ \\
\texttt{NMKV} & $12.721 \pm 0.706$ & $15.914 \pm 1.393$ & $12.681 \pm 0.143$ & $3.338 \pm 0.122$ & $14.270 \pm 0.038$ & $9.339 \pm 0.082$ \\
\texttt{Meta-FM} & $2.783 \pm 0.067$ & $1.770 \pm 0.186$ & $8.912 \pm 0.053$ & $0.538 \pm 0.020$ & $11.793 \pm 0.082$ & $4.809 \pm 0.119$ \\
\texttt{M2M-ED} & $8.422 \pm 0.244$ & $9.244 \pm 0.596$ & $11.783 \pm 0.442$ & $1.058 \pm 0.252$ & $21.925 \pm 0.494$ & $9.000 \pm 0.264$ \\
\texttt{M2M-OTMSE} & $4.504 \pm 0.572$ & $7.223 \pm 1.143$ & $10.859 \pm 0.166$ & $5.282 \pm 0.497$ & $14.802 \pm 0.225$ & $17.254 \pm 0.440$ \\
\texttt{M2M}-$\mathcal{W}_1$ & $3.237 \pm 0.304$ & $4.693 \pm 0.608$ & $10.251 \pm 0.108$ & $5.190 \pm 0.178$ & $15.111 \pm 0.169$ & $17.841 \pm 0.329$ \\
 \tfm& $\mathbf{1.934 \pm 0.018}$ & $\mathbf{1.388 \pm 0.020}$ & $\mathbf{8.323 \pm 0.054}$ & $\mathbf{0.330 \pm 0.015}$ & $\mathbf{10.438 \pm 0.047}$ & $\mathbf{4.136 \pm 0.102}$ \\ 
\bottomrule
\end{tabular}}
\label{tab:50_mkv_results}
\end{table}

\begin{table}[ht]
\centering
{\caption{\textbf{Prediction performance comparison on $\mathbf{100}$-dimensional McKean-Vlasov systems.} Mean$\pm$Std computed over $6$ random seeds. \textbf{Bold} indicates the best performing method.}}
\resizebox{\textwidth}{!}{%
\begin{tabular}{lcccccc}
\toprule
 & \multicolumn{2}{c}{Kuramoto} & \multicolumn{2}{c}{FitzHugh-Nagumo} & \multicolumn{2}{c}{Atlas} \\
Method & $\mathcal{W}_1$ & $ED$ & $\mathcal{W}_1$ & $ED$ & $\mathcal{W}_1$ & $ED$ \\
\midrule
\texttt{CFM} & $11.850 \pm 0.085$ & $7.215 \pm 0.105$ & $14.929 \pm 0.056$ & $0.651 \pm 0.008$ & $20.502 \pm 0.080$ & $6.729 \pm 0.040$ \\
\texttt{MF-Transformer} & $5.462 \pm 0.098$ & $3.315 \pm 0.202$ & $23.435 \pm 0.254$ & $3.305 \pm 0.276$ & $17.175 \pm 0.026$ & $8.721 \pm 0.026$ \\
\texttt{NMKV} & $23.441 \pm 1.559$ & $27.716 \pm 3.117$ & $22.191 \pm 0.109$ & $3.376 \pm 0.097$ & $21.133 \pm 0.077$ & $12.835 \pm 0.131$ \\
\texttt{Meta-FM} & $4.532 \pm 0.199$ & $3.258 \pm 0.320$ & $14.801 \pm 0.175$ & $1.028 \pm 0.053$ & $18.656 \pm 0.093$ & $7.641 \pm 0.042$ \\
\texttt{M2M-ED} & $14.400 \pm 0.065$ & $17.456 \pm 0.326$ & $22.320 \pm 0.390$ & $3.653 \pm 0.293$ & $32.217 \pm 0.268$ & $12.429 \pm 0.128$ \\
\texttt{M2M-OTMSE} & $6.751 \pm 1.154$ & $10.964 \pm 2.306$ & $17.311 \pm 0.297$ & $11.812 \pm 0.608$ & $21.407 \pm 0.367$ & $25.246 \pm 0.738$ \\
\texttt{M2M-}$\mathcal{W}_1$ & $5.204 \pm 0.436$ & $7.875 \pm 0.872$ & $14.757 \pm 0.077$ & $7.610 \pm 0.163$ & $20.358 \pm 0.044$ & $23.132 \pm 0.086$ \\
 \tfm & $\mathbf{3.005 \pm 0.037}$ & $\mathbf{2.299 \pm 0.092}$ & $\mathbf{12.687 \pm 0.084}$ & $\mathbf{0.548 \pm 0.033}$ & $\mathbf{15.940 \pm 0.055}$ & $\mathbf{6.633 \pm 0.130}$ \\
\bottomrule
\end{tabular}}
\label{tab:100_mkv_results}
\end{table}

\begin{table}[ht]
\centering
\caption{Ablation over the number of inference steps per marginal for the 100 dimensional McKean Vlasov systems}
\begin{tabular}{lcccccc}
\toprule
 & \multicolumn{2}{c}{Kuramoto 100D} \\
Method & $\mathcal{W}_1 (\downarrow)$ & $ED (\downarrow)$ \\
\midrule
\texttt{\tfm} (Steps=1) & $2.868 \pm 0.058$ & $2.630 \pm 0.116$ \\
\texttt{\tfm} (Steps=5) & $2.974 \pm 0.034$ &  $2.327 \pm 0.094$ \\
\texttt{\tfm} (Steps=10) & $2.992 \pm 0.035$ & $2.316 \pm 0.092$ \\
\texttt{\tfm} (Steps=20) & $2.998 \pm 0.037$ & $2.305 \pm 0.092$ \\
\texttt{\tfm} (Steps=50) & $3.003 \pm 0.037$ & $2.300 \pm 0.092$ \\
\texttt{\tfm} (Steps=100) & $3.005 \pm 0.037$ & $2.299 \pm 0.092$ \\
\bottomrule
\end{tabular}
\label{tab:steps_mkv}

\end{table}

\begin{table}[ht]
\centering
\caption{Cosine Distance between true and learned vector field for 2 dimensional systems averaged over 6 seeds.}
\begin{tabular}{lccc}
\toprule
 Method & \multicolumn{1}{c}{Atlas} & \multicolumn{1}{c}{FitzHugh-Nagumo} & \multicolumn{1}{c}{Kuramoto} \\
\midrule
\texttt{NMKV} & $1.436 \pm 0.004$ & $0.493 \pm 0.002$ & $0.557 \pm 0.010$ \\
\texttt{CFM} & $0.169 \pm 0.002$ & $0.098 \pm 0.004$ & $0.208 \pm 0.001$ \\
\texttt{Meta-FM} & $0.100 \pm 0.006$ & $0.074 \pm 0.005$ & $0.118 \pm 0.004$ \\
\tfm & $0.062 \pm 0.001$ & $0.070 \pm 0.001$ & $0.084 \pm 0.001$ \\
\bottomrule
\end{tabular}
\label{tab:cosine_dim2}
\end{table}

\begin{table}[ht]
\centering
\caption{Cosine Distance between true and learned vector field for 50 dimensional systems averaged over 6 seeds.}
\begin{tabular}{lccc}
\toprule
Method & \multicolumn{1}{c}{Atlas} & \multicolumn{1}{c}{FitzHugh-Nagumo} & \multicolumn{1}{c}{Kuramoto} \\
\midrule
\texttt{NMKV} & $1.248 \pm 0.003$ & $0.319 \pm 0.003$ & $0.528 \pm 0.003$ \\
\texttt{CFM} & $0.438 \pm 0.001$ & $0.087 \pm 0.001$ & $0.230 \pm 0.002$ \\
\texttt{Meta-FM} & $0.356 \pm 0.005$ & $0.102 \pm 0.003$ & $0.185 \pm 0.003$ \\
\tfm & $0.241 \pm 0.003$ & $0.048 \pm 0.002$ & $0.127 \pm 0.001$ \\
\bottomrule
\end{tabular}
\label{tab:cosine_dim50}
\end{table}

\begin{table}[ht]
\centering
\caption{Cosine Distance between true and learned vector field for 100 dimensional systems averaged over 6 seeds.}
\begin{tabular}{lccc}
\toprule
 Method & \multicolumn{1}{c}{Atlas} & \multicolumn{1}{c}{FitzHugh-Nagumo} & \multicolumn{1}{c}{Kuramoto} \\
\midrule
\texttt{NMKV} & $1.199 \pm 0.003$ & $0.422 \pm 0.002$ & $0.456 \pm 0.001$ \\
\texttt{CFM} & $0.480 \pm 0.002$ & $0.124 \pm 0.001$ & $0.252 \pm 0.002$ \\
\texttt{Meta-FM} & $0.476 \pm 0.002$ & $0.182 \pm 0.010$ & $0.215 \pm 0.003$ \\
\tfm & $0.299 \pm 0.003$ & $0.071 \pm 0.003$ & $0.149 \pm 0.001$ \\
\bottomrule
\end{tabular}
\label{tab:cosine_dim100}
\end{table}

\begin{table}[ht]
\centering
{\caption{\textbf{Prediction performance comparison on two simple $\mathbf{2}$-dimensional McKean-Vlasov systems (GoToMean and GoToZero).} Mean$\pm$Std computed over $3$ random seeds. \textbf{Bold} indicates the best or tied for best performing method.}}
\resizebox{\textwidth}{!}{%
\begin{tabular}{lcccc}
\toprule
 & \multicolumn{2}{c}{GoToZero} & \multicolumn{2}{c}{GoToMean} \\
Method & $\mathcal{W}_1$ & $ED$ & $\mathcal{W}_1$ & $ED$ \\
\midrule
\texttt{CFM} &  $\mathbf{0.014 \pm 0.001} $ & $0.003 \pm 0.001$ & $1.782 \pm 0.018$ & $1.162 \pm 0.016$ \\
\texttt{MF-Transformer} & $0.127 \pm 0.027$ & $0.166 \pm 0.050$ & $2.059 \pm 0.490$ & $2.350 \pm 0.836$ \\
\texttt{NMKV}& $0.394 \pm 0.000$ & $0.218 \pm 0.001$ & $1.142 \pm 0.099$ & $1.082 \pm 0.166$ \\
\texttt{Meta-FM} & $0.026 \pm 0.004$ & $0.022 \pm 0.007$ & $0.563 \pm 0.052$ & $0.414 \pm 0.057$ \\
\texttt{M2M-MSE}  & $\mathbf{0.014 \pm 0.000}$ & $\mathbf{0.002 \pm 0.000}$ & $0.851 \pm 0.165$ & $0.815 \pm 0.090$  \\
\texttt{M2M-}$\mathcal{W}_1$ & $0.015 \pm 0.002$ & $\mathbf{0.002 \pm 0.001}$ & $\mathbf{0.355 \pm 0.003}$ & $\mathbf{0.372 \pm 0.008}$ \\
\tfm & $0.016 \pm 0.001$ & $0.005 \pm 0.001$ & $0.465 \pm 0.044$ & $0.555 \pm 0.076$ \\
\bottomrule
\end{tabular}}
\label{tab:extra_dim2}
\end{table}

\begin{table}[ht]
\centering
{\caption{\textbf{Prediction performance comparison on two simple $\mathbf{50}$-dimensional McKean-Vlasov systems (GoToMean and GoToZero).} Mean$\pm$Std computed over $3$ random seeds. \textbf{Bold} indicates the best performing method.}}
\resizebox{\textwidth}{!}{%
\begin{tabular}{lcccc}
\toprule
 & \multicolumn{2}{c}{GoToZero} & \multicolumn{2}{c}{GoToMean} \\
Method & $\mathcal{W}_1$ & $ED$ & $\mathcal{W}_1$ & $ED$ \\
\midrule
\texttt{CFM}  & $0.242 \pm 0.002$ & $\mathbf{0.014 \pm 0.001}$ & $10.727 \pm 0.086$ & $6.981 \pm 0.222$ \\
\texttt{MF-Transformer}  & $2.467 \pm 0.015$ & $0.900 \pm 0.010$ & $26.676 \pm 0.554$ & $11.834 \pm 1.191$ \\
\texttt{NMKV}& $7.441 \pm 0.298$ & $4.475 \pm 0.225$ & $9.700 \pm 0.720$ & $6.919 \pm 1.213$ \\
\texttt{Meta-FM} & $0.287 \pm 0.007$ & $0.104 \pm 0.011$ & $6.056 \pm 0.215$ & $3.518 \pm 0.212$ \\
\texttt{M2M-MSE} & $0.272 \pm 0.000$ & $0.116 \pm 0.002$ & $12.942 \pm 1.270$ & $21.576 \pm 2.543$ \\
\texttt{M2M-}$\mathcal{W}_1$ & $\mathbf{0.238 \pm 0.002}$ & $0.062 \pm 0.006$ & $5.267 \pm 0.424$ & $6.273 \pm 0.834$ \\
\tfm & $0.266 \pm 0.011$ & $0.043 \pm 0.008$ & $\mathbf{3.804 \pm 0.010}$ & $\mathbf{2.149 \pm 0.065}$ \\
\bottomrule
\end{tabular}}
\label{tab:extra_dim50}
\end{table}

\begin{table}[ht]
\centering
{\caption{\textbf{Prediction performance comparison on two simple $\mathbf{100}$-dimensional McKean-Vlasov systems (GoToMean and GoToZero).} Mean$\pm$Std computed over $3$ random seeds. \textbf{Bold} indicates the best performing method.}}
\resizebox{\textwidth}{!}{%
\begin{tabular}{lcccc}
\toprule
 & \multicolumn{2}{c}{GoToZero} & \multicolumn{2}{c}{GoToMean} \\
Method & $\mathcal{W}_1$ & $ED$ & $\mathcal{W}_1$ & $ED$ \\
\midrule
\texttt{CFM}& $0.548 \pm 0.003$ & $\mathbf{0.070 \pm 0.002}$ & $21.315 \pm 0.241$ & $11.518 \pm 0.226$ \\
\texttt{MF-Transformer}& $3.687 \pm 0.034$ & $1.340 \pm 0.019$ & $40.543 \pm 0.603$ & $19.544 \pm 1.521$ \\
\texttt{NMKV} & $17.954 \pm 0.159$ & $11.714 \pm 0.145$ & $28.114 \pm 0.499$ & $16.447 \pm 0.742$ \\
\texttt{Meta-FM} & $0.417 \pm 0.008$ & $0.121 \pm 0.013$ & $10.136 \pm 0.084$ & $5.920 \pm 0.178$ \\
\texttt{M2M-MSE} & $0.388 \pm 0.003$ & $0.161 \pm 0.006$ & $19.104 \pm 1.470$ & $32.062 \pm 2.919$ \\
\texttt{M2M-}$\mathcal{W}_1$  & $\mathbf{0.348 \pm 0.001}$ & $0.103 \pm 0.002$ & $6.520 \pm 0.578$ & $6.899 \pm 1.155$ \\
\tfm & $0.439 \pm 0.068$ & $0.085 \pm 0.036$ & $\mathbf{5.535 \pm 0.105}$ & $\mathbf{2.761 \pm 0.182}$ \\
\bottomrule
\bottomrule
\end{tabular}}
\label{tab:extra_dim100}
\end{table}

\end{document}